\documentclass[runningheads]{llncs}
\usepackage{graphicx}
\usepackage{amsmath,amssymb} 
\usepackage{color}
\usepackage{comment}

\usepackage[width=122mm,left=12mm,paperwidth=146mm,height=193mm,top=12mm,paperheight=217mm]{geometry}

\makeatother
\renewcommand\labelitemii{$\m@th\bullet$}
\makeatletter

\begin{document}
\pagestyle{headings}
\mainmatter
\def\ECCV18SubNumber{2577}  

\newcommand{\gs}[1]{\textcolor{red}{\textbf{#1}}}

\title{Task-Driven Super Resolution:\\
Object Detection in Low-resolution Images} 

\titlerunning{Task-Driven Super Resolution}
\authorrunning{M. Haris et al.}

\author{Muhammad Haris$^1$, Greg Shakhnarovich$^2$, and Norimichi Ukita$^1$}
\institute{$^1$TTI, Japan $^2$TTI-C, United States\\
{\small \{mharis, ukita\}@toyota-ti.ac.jp, greg@ttic.edu}}

\maketitle

\begin{abstract}
  We consider how image super resolution (SR) can contribute to an
  object detection task in low-resolution images. Intuitively, SR
  gives a positive impact on the object detection task. While several
  previous works demonstrated that this intuition is correct, SR and
  detector are optimized independently in these works. This paper
  proposes a novel framework to train a deep neural network where the
  SR sub-network explicitly incorporates a detection loss in its
  training objective, via a tradeoff with a traditional detection
  loss.  This end-to-end training procedure allows us to train SR
  preprocessing for any differentiable detector. We demonstrate that
  our task-driven SR consistently and significantly improves accuracy
  of an object detector on low-resolution images for a variety of
  conditions and scaling factors.
  
\end{abstract}

\section{Introduction}
\label{section:introduction}

Image Super Resolution (SR) belongs to image restoration and
enhancement (e.g., denoising and deblurring) algorithms,
widely studied in computer vision and graphics.  In both
communities, the goal is to reconstruct an image from a degenerated
version as accurately as possible.
The quality of the reconstructed image is evaluated by pixel-based
quantitative metrics such as PSNR (peak signal-to-noise ratio) and
SSIM (structure similarity) \cite{DBLP:journals/tip/WangBSS04}.
Recently-proposed perceptual quality
\cite{DBLP:conf/eccv/JohnsonAF16,DBLP:conf/nips/DosovitskiyB16,DBLP:conf/iccv/SajjadiSH17}
can be also employed for evaluation as well as for optimizing the
reconstruction model.
Relationships between the pixel-based and perceptual quality metrics
have been investigated in the literature
\cite{DBLP:conf/icdsp/HanhartKE13,DBLP:conf/icip/KunduE15} in order to
harmonize these two kinds of metrics. Ultimately, the goal of SR is
still to restore an image as well as possible in accordance 
with criteria in human visual perception.

One connection between SR, and other image restoration tools, and visual recognition is that
despite continuing advances in visual recognition, it remains
vulnerable to a wide range of image degradation, including low
resolution and
blur~\cite{vasiljevic2016examining,dodge2016understanding}. Image
restoration such as SR can
serve as an input enhancement step to alleviate this vulnerability.
For example, accuracy of many recognition tasks can be improved by deblurring
\cite{DBLP:journals/pami/NishiyamaHTSKY11,DBLP:conf/cvpr/ChenHYW11,DBLP:conf/bmvc/HradisKZS15,DBLP:conf/eccv/XiaoWHH16}
or denoising \cite{DBLP:conf/icip/MilaniBR12}.
SR has been also shown to be effective for such preprocessing for
several recognition
tasks~\cite{DBLP:conf/wacv/DaiWCG16,DBLP:conf/cvpr/Hennings-YeomansBK08,DBLP:conf/icip/Hennings-YeomansKB09,DBLP:conf/icb/ShekharPC11,DBLP:conf/bmvc/BilgazyevESK11}.

\begin{figure}[!t]
  \begin{center}
    \begin{tabular}[c]{ccccc}
      \includegraphics[width=.18\textwidth]{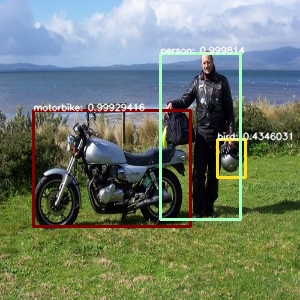}
      &
        \includegraphics[width=.18\textwidth]{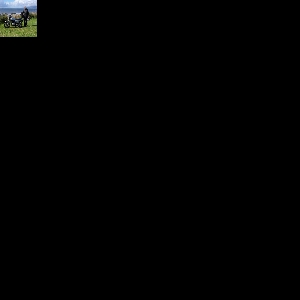}
      &
        \includegraphics[width=.18\textwidth]{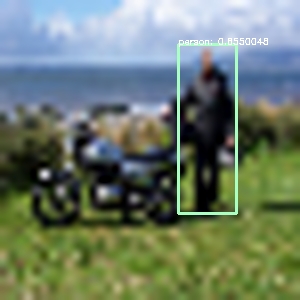}
      &
        \includegraphics[width=.18\textwidth]{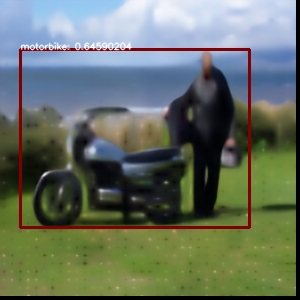}
      &
        \includegraphics[width=.18\textwidth]{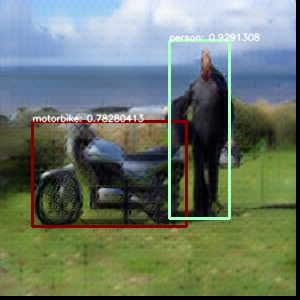}\\
      {\scriptsize (a) HR}
      &{\scriptsize (b) LR}
      &{\scriptsize (c) Bicubic SR}
      &{\scriptsize (d) SR (no task)}
      &{\scriptsize (e) Task driven SR\vspace{-.5em}}\\
      &&&&{\scriptsize (proposed)}\\
      &&{\scriptsize PSNR: 21.26} & {\scriptsize PSNR: 22.02 }& {\scriptsize PSNR: 21.54}
    \end{tabular}
    
    \caption{Scale sensitivity in object recognition and the
      effectiveness of our proposed method (i.e., end-to-end learning
      in accordance with an object recognition task). Images shown in
      the top row show (a) an original high resolution image, (b) its
      low-resolution image (here $1/8$-size, padded with black), (c) SR image
      obtained by bicubic interpolation, (d) SR image obtained by the SR model
      optimized with no regard to detection, and (e) SR image
      obtained by our proposed task-driven SR method, using the same model as in
      (d). For each of the reconstructed HR images, we also report
      PSNR w.r.t. the original. Despite ostensibly lower PSNR, the TDSR result recovers the correct
      detection results with high scores, in this case even
      suppressing a false detection present in the original HR input,
      and at the same produces a plausible looking HR image.
    }
    \label{fig:contribution}
  \end{center}
\end{figure}

Typically, in such applications the SR is trained in isolation from
the downstream task, with the only connection through the selection of
images to train or fine-tune the SR method (e.g., for character
recognition, SR is trained on character images).

We propose to bridge this divide, and explicitly incorporate the
objective of the downstream task (such as object detection) into
training of an SR module.
Figure \ref{fig:contribution} illustrates the effect of our proposed,
task-driven approach to SR. SR images (c),
(d), and (e) generated from a low-resolution (LR) image (b)
can successfully bring recognition accuracy close to the score of
their original high-resolution (HR) image (a).
%
%


%
Our approach is motivated by two observations:
\begin{description}
\item[SR is ill-posed:] many HR images when downsampled produce the
  same LR image. We expect that the additional cue given by the
  downstream task objective such as detection may help guide the SR
  solution.
\item[Human perception and machine perception differ:] It is known that big
  differences are observed between human and machine perceptions, in
  particular, with highly-complex deep networks. This is perhaps best
  exemplified by adversarial
  images~\cite{szegedy2013intriguing,DBLP:conf/cvpr/NguyenYC15,DBLP:conf/cvpr/Moosavi-Dezfooli16}
  that can ``fool'' machine perception but not human. Thus, if our
  goal is to super-resolve an image for machine
  consumption, we believe it is prudent to explicitly ``cater'' to the
  machine perception when learning SR.
\end{description}

The two SR images in Fig~\ref{fig:contribution} (d) and (e) illustrate these
points. Both look similar to the human eye, but the detection results
by a network
differ significantly. Furthermore, the conventional measure of
reconstruction quality (PSNR) fails to capture the difference,
assigning significantly higher value to (d) which yields to much
worse detection results.
The main contributions of this paper are:

\begin{itemize}
\item An approach to super-resolution that uses the power of end-to-end training in deep
  learning to combine low-level and high-level vision objectives,
  leading to what we call {\em Task-Driven Super Resolution}
  (TDSR). As a means of increasing robustness of object detection to
  low resolution inputs, this approach provides results substantially better than
  other SR methods, and is 
  potentially applicable to a broad range of low-level image
  processing tools and high-level tasks.

\item A novel view of super-resolution, explicitly acknowledging the
  generative or semantic aspects of SR in high scaling factors, which we
  hope will encourage additional work in the community to help further reduce
  the gap between low-level and high-level vision.
\end{itemize}

\section{Related Work}
\label{section:related}
While there has been much work on super-resolution and on evaluating
and improving some measure of perceptual quality of images,
comparatively little
work exist on optimizing image restoration tools for machine perception.

\subsubsection{Image quality assessment}
\label{subsection:vqa}

Image restoration and enhancement require appropriate quality
assessment metrics both for evaluation and (when machine
learning is used) as training objectives. As mentioned in
Sec.~\ref{section:introduction}, PSNR and SSIM
\cite{DBLP:journals/tip/WangBSS04} are widely used as such metrics, focusing
on comparing a reconstructed/estimated image with its ground
truth image. There exist methods for quality assessment that do not
require a reference ground truth
image~\cite{DBLP:conf/icip/Luo04,DBLP:conf/cvpr/TangJK14}, including
some that use deep neural networks to learn the metrics~\cite{DBLP:conf/cvpr/KangYLD14,DBLP:journals/tip/MaLZDWZ18}.
%

Several quality assessment
metrics~\cite{DBLP:conf/icip/ReibmanBG06,DBLP:conf/icip/YeganehRW12,DBLP:conf/icip/FangLZLG16}
have been evaluated specifically for SR, including no-reference
metrics~\cite{DBLP:journals/cviu/MaYY017}. However all of these
metrics are a proxy for (assumed or approximated) human judgment
perceptual quality, and do not consider high-level visual tasks such
as recognition.

Some task-dependent quality assessment metrics have been proposed for
certain tasks, including biometrics~\cite{DBLP:journals/tip/GalballyMF14}, face recognition~\cite{DBLP:conf/icip/PulecioBB17}, and object recognition
\cite{plosone/Yuan17}, showing improvements vs. the task-agnostic
metrics. None of them, however, have been used in a joint learning
framework with the underlying image enhancement such as SR.

\subsubsection{Image Super Resolution}
\label{subsection:isr}
A huge variety of image SR techniques have been proposed; see survey
papers
\cite{DBLP:journals/ivc/Ouwerkerk06,DBLP:journals/mva/NasrollahiM14,DBLP:conf/eccv/YangMY14}
for more details.
While self-contained SR is attractive (e.g., self-similarity based SR
\cite{DBLP:conf/accv/YangHY10,DBLP:conf/iccv/MichaeliI13,DBLP:conf/cvpr/HuangSA15}),
most recent SR algorithms utilize external training images for higher
performance; for example, exemplar based
\cite{DBLP:conf/cvpr/TimofteRG16,DBLP:journals/tip/YangWLCH12,iclr/Bansal18},
regression based
\cite{DBLP:journals/pami/KimK10,DBLP:conf/cvpr/Perez-Pellitero16},
and web-retrieval based \cite{DBLP:journals/tip/YueSYW13}.
The effectiveness of using both self and external images is explored in~\cite{DBLP:conf/cvpr/YangLC13,DBLP:journals/tip/WangYWCYH15}.

Like other vision problems, SR has benefited from recent advances in
deep convolutional neural networks (DCNNs).
SRCNN \cite{dong2016image} enhances the spatial resolution of an input
LR image by hand-crafted upsampling filters. The enlarged image is
then improved by a DCNN. Further improvements are achieved with more
advanced architectures, introducing residual
connections~\cite{Kim_2016_VDSR,szegedy2015going} and recursive layers
\cite{kim2016deeply}, however the use of the hand-crafted upsampling filters
remains an impediment. That can be alleviated by embedding an upsampling layer into
a DCNN
\cite{dong2016accelerating,shi2016real,Lim_2017_CVPR_Workshops}.
Progressive upsampling \cite{cvpr/LapSRN17} is also effective for
leveraging information from different scales.
By sharing the SR features at different scales by iterative forward
and backward projections, DBPN-SR \cite{arxiv/Haris18} enables the networks
to preserve the HR components by learning various up- and
down-sampling operators while generating deeper features.

While deep features provided by DCNNs allow us to preserve clear
high-frequency photo-realistic textures, it is difficult to completely
eliminate blur artifacts. This problem has been addressed by
introduction of novel objectives, such as perceptual
similarity~\cite{DBLP:conf/eccv/JohnsonAF16,DBLP:conf/nips/DosovitskiyB16}
and adversarial losses~\cite{goodfellow2014generative,DBLP:conf/eccv/YuP16}.
Finally, the two ideas can be combined, incorporating perceptual
similarity into generative adversarial networks
(GANs) in SRGAN~\cite{ledig2016photo}.

In contrast to prior work, we explicitly incorporate the objective of a well defined,
discriminative task (such as detection) into the SR framework.


\subsubsection{Object detection}\label{sec:object_det}
Most state-of-the-art object detection algorithms extract or
evaluate object proposals (e.g., bounding boxes)
\cite{DBLP:journals/ijcv/UijlingsSGS13,DBLP:conf/eccv/ZitnickD14,DBLP:conf/iccv/Girshick15,DBLP:journals/pami/HosangBDS16,DBLP:journals/pami/GirshickDDM16} within a query image and
evaluate the ``objectness'' of each bounding box for object
detection, using DCNN features computed or pooled over each box. In many recent models, the mechanism for producing
candidate boxes is incorporated into the network architecture~\cite{DBLP:journals/pami/RenHG017,DBLP:conf/iccv/HeGDG17}.

Unlike approaches using object proposals, SSD~\cite{DBLP:conf/eccv/LiuAESRFB16} and YOLO9000~\cite{DBLP:conf/cvpr/RedmonF17} use pre-set default boxes
(a.k.a. anchor boxes) covering a query image.  The objectness score is
computed for each object category in all boxes while its spatial
parameters (e.g., location, scale, and aspect ratio) are
optimized. This streamlines the computation at test time and produces extremely
fast, as well as accurate, detection framework.

\subsubsection{Cnnnections to generative models}\label{sec:gen}
There is also an interesting connection between our approach and the
gradient-based adversarial images~\cite{szegedy2013intriguing} as well as the popular ``neural art''
technique called DeepDream~\cite{deepdream}. In both of those,
an input image (at full resolution) is modified using gradient
descent with the objective to achieve certain output for an image
classification network. For adversarial images the goal is to make the
network predict an incorrect class, while in DeepDream the goals are
aesthetic.

\section{Task Driven Super-resolution}\label{sec:method}

Our method relies on two building blocks: a super-resolution (SR)
network $S$ and a task network $D$. The
SR network maps a low-resolution image $x^l$ to a high-resolution
image $x^h=S(x^l;\theta_{SR})$, where $\theta_{SR}$ denotes all the
parameters of the network. The task network takes an image $x$ and
outputs a (possibly structured) prediction
$\widehat{y}=D(x;\theta_{D})$. We refer to these predictors as
``networks'' because they are currently likely to be deep
neural networks. However our approach does not presume anything about
$S$ and $D$ beyond
differentiability.

We assume that the task network $D$ has been trained and its
parameters $\theta_D$ remain fixed
throughout (and will, for brevity, be omitted from notation). Thus, our method is applicable to any task network, and can be
used to make an off-the-shelf network that fails on low resolution
inputs more robust to such inputs. It can be used for a variety of tasks, for example, depth estimation or semantic
segmentation. However in this paper we restrict our attention to the object
detection task, in which $\widehat{y}$ consists of a
set of scored bounding boxes for given object classes.

\subsection{Component networks}\label{sec:components}
We use the recently proposed Deep Back-Projection Networks (DBPN)~\cite{arxiv/Haris18} as
the SR component. The DBPN achieve state of the art or competitive results on
standard SR benchmarks, when trained with the
MSE reconstruction loss
\begin{equation}
  \label{eq:Lrec}
  L_{rec}\left(x,\widehat{x})\right)\,=\,\frac{1}{N}\sum_{i=1}^N(x_i-\widehat{x}_i)^2
\end{equation}
where $i$ ranges of the $N$ pixel indices in the HR image $x$.

As the detector, we use the Single Shot MultiBox Detector
(SSD)~\cite{DBLP:conf/eccv/LiuAESRFB16}. The SSD detector works with a
set of default bounding boxes, covering a range of positions, scales
and aspect ratios; each box is scored for presence of an object from
every class. Given the ground truth for an image $x$, a
subset of $B$ default boxes is matched to the ground truth boxes, this
matches forming the predicted detections $\widehat{y}(x)$. The
task (detection) loss of SSD is combined of
of confidence loss and localization loss: 
\begin{equation}
 \label{eq:Ltask}
 L_{task}(y,\widehat{y}(x))\,=\,\frac{1}{B}\left[
   L_{conf}(y,\widehat{y}(x))\,+\,\lambda L_{loc}(y,\widehat{y}(x))
 \right].
\end{equation}
The confidence loss $L_{conf}$ penalizes incorrect class
predictions for the matched boxes. The localization loss $L_{loc}$
penalizes displacement of boxes vs. the ground truth, using smooth
$L_1$ distance. Both losses in~\eqref{eq:Ltask} are differentiable with respect to their inputs.

Importantly, every default bounding box in SSD is associated with a
set of cells in feature maps (activation layers) computed by a convolutional neural
network. As a result, since the loss in~\eqref{eq:Ltask} decomposes
over boxes, it is a differentiable function of the network activations and thus a
function of the pixels in the input image, allowing us to incorporate
this task loss in the TDSR objective described below.

Both of our chosen component networks have code made publicly
available by their authors, and can be trained end to end, providing a
convenient testbed for our approach; many other choices are possible,
in particular for the detector component, but we do not explore them
in this paper.

\subsection{Task driven training}\label{sec:task}
Normally, learning-based SR systems are trained using some sort of
reconstruction loss $L_{rec}$, such as mean (over pixels) squared error (MSE)
between $x$ and the downsampled version of $x$ superresolved by
$S$. In contrast, the detector is trained with a surrogate loss $L_{task}$
intended to improve the measure of its accuracy, typically measured as
the average precision (AP) for one class, and the mean AP (mAP) over
classes for the entire data set/task.

Let $x$ be the image from the detection data set, with detection
ground truth labels $y$, and let $\downarrow(\cdot)$ denote downscaling of an
image by a fixed factor. We propose the
compound loss, which on the example $(x,y)$ is given by
\begin{equation}
  \label{eq:losses}
  L(x,y;\theta_{SR})\,=\,
  \alpha L_{rec}\left(x,S(\downarrow(x);\theta_{SR})\right)\,+\,
  \beta L_{task}\left(y,D(S(\downarrow(x);\theta_{SR}))\right),
\end{equation}
where $\alpha$ and $\beta$ are weights determining relative strength
of the fidelity term (reconstruction loss) and the semantic term
(detection loss). Under the assumption that both $S$ and $D$ are
differentiable, we can use the chain rule, and compute the gradient of
$L_{D}$ with respect to its input, the super-resolved
$\downarrow(x)$. Then this per-pixel gradient is combined with the
per-pixel gradient of the reconstruction loss $L_{rec}$:
The SR
parameters $\theta_{SR}$ are then updated using
standard back-propagation from this combined gradient:

\begin{equation}
  \label{eq:grad}
  \alpha\frac{\partial }{\partial\theta_{SR}}
  L_{rec}\left(x,S(\downarrow(x);\theta_{SR})\right)
  +
  \beta\frac{\partial L_{task}\left(y,D(S(\downarrow(x)))\right)}{\partial S(\downarrow(x))}
  \frac{\partial S(\downarrow(x))}{\partial \theta_{SR}}
\end{equation}

\subsubsection{Interpretation}\label{sec:interpretation}
 As mentioned in
Section~\ref{section:introduction}, SR is an ill-posed problem. At
sufficiently high upscaling factors, it resembles (conditional) image
\emph{generation} more than image \emph{restoration}, since
a large
amount of information destroyed in downscaling process must
effectively be ``hallucinated''. Most current image generation
methods, such based on GANs or autoencoders, either do not explicitly
regard the semantic content of the generated image, or ``hardcode'' it
into the generator by training only on images of a specific class. Our
objective~\eqref{eq:losses} encourages the image to both look good to
a human
(similar to the original) and look correct to the machine (yield the
same recognition results). The values of $\alpha$ and $\beta$ control
this tradeoff. With $\alpha \gg \beta$, we effectively ignore the
downstream task, and get the traditional, MSE-driven SR learning, with
the limitations for downstream detection discussed in
Section~\ref{section:related} and demonstrated in Section~\ref{section:experimental_results}.

With
$\beta \gg \alpha$ we effectively ignore the original high resolution image, and
the objective is purely semantic. In this case, intuitively, if the
``SR'' method were to simply paste a fixed
canonical object of the correct class at the appropriate location and
scale in the image, and the detector correctly picks up on these
objects, we get a perfect value of the task loss. However, in this
hypothetical scenario we would in effect replace the SR with a
LR detector. That of course would bring up back to the original challenges
of LR detection. We also would not get the extra benefit of creating
human-interpretable intermediate HR image, connected to the original
LR input. 

We expect the optimal tradeoff to be somewhere between these
scenarios, incorporating meaningful contributions
from both the reconstruction and the detection objectives. The precise
``mixing'' of the two is subject to algorithm design, as detailed below.

\subsection{Training schedules}\label{sec:schedule}
The definition of loss in~\eqref{eq:losses} depends on the values of
$\alpha$ and $\beta$, and we can consider a number of settings, both
static (fixed weights) and dynamic (weights changing through
training). We describe these here, and evaluate them in Section~\ref{section:experimental_results}.

\noindent\textbf{Fine-tune} Generally, we assume that $S$ has been trained for
super-resolution for a given factor on images from a domain that could
be different from the domain of $D$. We can simply fine-tune SR on the
new domain, without incorporating the task loss: $\alpha=1$,
$\beta=0$.

\noindent\textbf{Balanced} We can start with a phase of fine-tuning
the SR on reconstruction only ($\alpha=1$, $\beta=0$) and then increase $\beta$ to a non-zero
value, introducing task-driven component. Note that the appropriate relative magnitude
of $\beta$ with respect to $\alpha$ will depend not only on the
desired tradeoff between the objectives, but also on the relative
scale of the two loss functions.

\noindent\textbf{Task only} Alternatively, we can forgo the
reconstruction driven phase, and fine-tune $S$ with task loss only,
$\alpha=0$, $\beta=1$.

\noindent\textbf{Gradual} Finally, we can gradually increase $\beta$,
from zero to a high value, training with each value for a number of
iterations. We could expect this schedule to provide a more gentle
introduction of the task objective, gradually refining the initially
purely reconstruction-driven SR.


\section{Experimental Results}
\label{section:experimental_results}
\subsection{Implementation Details}
\noindent{\textbf{Base networks}} 
DBPN~\cite{arxiv/Haris18} constructs mutually-connected up- and down-sampling layers each of which represents different types of image degradation and high-resolution components. The stack of up- and down- projection units creates an efficient way to iteratively minimize the reconstruction error, to reconstruct a huge variety of super-resolution features, and to enable large scaling factors such as $8\times$ enlargement. We used the setting recommended by the authors:
``a $8 \times 8$ convolutional layer with four striding and two padding''
and 
``a $12\times 12$ convolutional layer with eight striding and two padding''
are used for $4\times$ and $8\times$ SRs, respectively, in order to construct a projection unit.
For object detection, we use SSD300 where the input size is $300\times 300$ pixels. The network uses VGG16 through conv5\_3 layer, then uses conv4\_3, conv7 (fc7), conv8\_2, conv9\_2, conv10\_2, and conv11\_2 as feature maps to predict the location and confidence score of each detected object. The code for both networks are publicly accessible in the internet.


\noindent{\textbf{Datasets}} We initialized all experiments with DBPN
model pretrained on the DIV2K data set~\cite{Agustsson_2017_CVPR_Workshops},
made available by the authors of~\cite{arxiv/Haris18}.
We used SSD network pretrained on PASCAL VOC0712 trainval and publicly
available as well. 
When fine-tuning DBPN in our experiments, with or without task-driven
objective, we reused PASCAL VOC0712 trainval, with data augmentation. The augmentation consists of photometric distortion, scaling, flipping, random cropping that are recommended to train SSD.
Test images on VOC2007 were used for testing in all experiments.
The input of DBPN was a LR image that was obtained by bicubic
downscaling the original (HR, $300\times300$) image from the data set
with a particular scaling factor (i.e., $1/4$ or $1/8$ in our
experiments, corresponding to $4\times$ and $8\times$ SR).

\noindent{\textbf{Training setting}} We used a batch size of 6. The
learning rate was initialized to $1e-4$ for all layers and decreased
by a factor of 10 for every $10^5$ iterations in the total $2 \times
10^5$ iterations for training runs consisting of 300,000 iterations.
For optimization, we used Adam with momentum set to $0.9$. All experiments were conducted using PyTorch on NVIDIA TITAN X GPUs.

\subsection{Comparison of Training Schedules}
Following the discussion in Sec.~\ref{sec:schedule}, we investigate
different settings and schedules for values of $\alpha$ and $\beta$
that control the reconstruction-detection tradeoff in~\eqref{eq:losses}.
Table~\ref{tab:acc_loss} shows PSNR and mAP for a number of schedules
described on the left in ($n$:$\alpha$:$\beta$) format, indicating
training for $n$ iterations with the corresponding values of $\alpha$
(weight on reconstruction loss) and $\beta$ (weight on detection
loss); $+$ indicates continuation of training. The schedules are (a)
SR: baseline using pretrained SR not fine-tuned on Pascal; (b) SR-FT:
fine-tuned for 100k iterations; (c) SR-FT+: fine-tuned for 300k
iterations; (d) TDSR-0.1: balanced schedule in which after 100k of
reconstruction-only training, we introduce detection loss with the
constant weight of $\beta=0.1$; (e) TDSR-0.01: same but the
$\beta=0.01$; (f) TDSR-DET: $\alpha=0$ so only detection (AP) loss is
used to fine-tune SR for 300k iterations; and finally (g) TDSR-grad:
gradual increase of $\beta$ to 1 throughout the 300k iterations.

\begin{figure}[!t]
  \begin{center}
    \begin{tabular}[c]{ccccc}
      \includegraphics[width=.18\textwidth]{images_temp/1012/1012_hr_original}
      &
        \includegraphics[width=.18\textwidth]{images_temp/1012/1012_8x_ftplus_original}
      &
        \includegraphics[width=.18\textwidth]{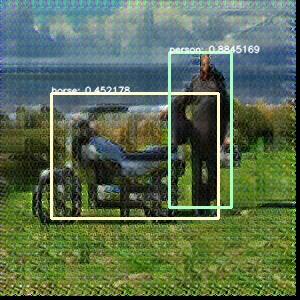}
      &
        \includegraphics[width=.18\textwidth]{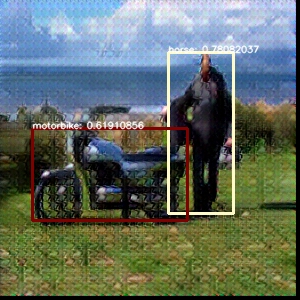}
      &
        \includegraphics[width=.18\textwidth]{images_temp/1012/1012_8x_balance_original}\\
      {\scriptsize (a) HR}
      &{\scriptsize (b) SR-FT+}
      &{\scriptsize (c) TDSR-DET}
      &{\scriptsize (d) TDSR-Grad}
      &{\scriptsize (e) TDSR-0.01\vspace{-.5em}}\\
      &{\scriptsize PSNR: 22.02 dB} &{\scriptsize PSNR: 16.63 dB} & {\scriptsize PSNR: 19.45 dB}& {\scriptsize PSNR: 21.54 dB}
    \end{tabular}
    
    \caption{Comparison on training schedules on $8\times$. PSNR
      values are for this image only.}
    \label{fig:figure2}
  \end{center}
\end{figure}

The values in the table provide us with multiple observations. First,
it helps to fine-tune SR on the new domain, so SR-FT has much higher
PSNR and mAP than SR. It helps to fine-tune for longer, hence better
results with SR-FT+ (in both PSNR and mAP), but we start observing
diminishing returns. Switching to variants of TDSR, we see a dramatic
increase in mAP accuracy. As the relative value of $\beta$ becomes
larger, we get additional improvements, but at the cost of
a significant decline in PSNR (and as we see in Fig.~\ref{fig:figure2}
and in
Section~\ref{sec:qual}, in visual quality). However, for a certain
regime, namely TDSR-0.01, we see a much higher mAP than the no-task
values, with only a marginal decline in PSNR. We thus identify this
schedule as the best based on our experiments. Finally, the numbers in the table
further illustrate that higher PSNR must not correspond to better
detection results.

\begin{table}[t]
\caption{Comparison of training schedules for~\eqref{eq:losses},
  evaluated on VOC2007 test. $n:\alpha:\beta$ indicates training for
  $n$ iterations with the given $\alpha$,$\beta$ values. See text for additional explanations.
The best score in each column is colored by \textcolor{red}{red}.
}
\centering
\label{tab:acc_loss}
\begin{tabular}{l|*1l|*2c|*2c}
\hline\noalign{\smallskip}
&HR: 75.78\% mAP &\multicolumn{2}{c}{$4\times$} & \multicolumn{2}{c}{$8\times$} \\         
Setting &$n$-iter : $\alpha : \beta$& PSNR &mAP& PSNR &mAP\\
\noalign{\smallskip}\hline\noalign{\smallskip}
SR &0k:1:0&22.80&41.9&17.50&10.6\\
SR-FT& 100k:1:0&26.60&52.6&22.70&22.0\\
SR-FT+& 100k:1:0+200k:1:0&\textcolor{red}{26.67}&53.6&\textcolor{red}{22.81}&22.9\\
TDSR-0.1& 100k:1:0+200k:1:0.1&25.13&61.6&21.08&36.1\\
TDSR-0.01& 100k:1:0+200k:1:0.01&24.01&\textcolor{red}{62.2}&22.24&\textcolor{red}{37.5}\\
TDSR-DET& 300k:0:1 &17.02&61.0&16.72&37.4\\
TDSR-Grad& 100k:1:0+70k:1:0.01+70k:1:0.1+60k:1:1&21.80&61.5&19.78&37.2\\
\noalign{\smallskip}\hline
\end{tabular}
\end{table}

%

Table~\ref{tab:natural} shows detailed results for comparing our TDSR
method to
other SR approaches, including the baseline bicubic SR, and
a recently proposed state-of-the-art SR method
(SRGAN~\cite{ledig2016photo}). Comparison to SRGAN is particularly
interesting since it uses a different kind of objective
(adversarial/perceptual) which may be assumed to be better suited for
task-driven SR.
Note that all the other SR models were just pretrained, and not
fine-tuned on Pascal. 
We also compared results obtained directly from LR images (padded with
black to fit to the pretrained SSD300 detector).

We see that reduction in resolution has a drastic effect on the mAP of
the detector, dropping it from 75.8 to 41.7 for $4\times$ and 16.6 for
$8\times$. This is presumably due to both the actual loss of
information, and the limitations of the detector architecture which
may miss small bounding boxes. The performance is not significantly
improved by non-task-driven SR methods, which in some cases actually
harm it further! 
%
However, our proposed TDSR approach obtains significantly better
results for both scaling factors, and recovers a significant fraction of the detection accuracy
lost in LR. 

\begin{table}[t]
\caption{VOC2007 test detection results on $4\times$ and $8\times$ enlargement. Note: Original images (HR) obtained 75.8\% mAP}
\centering
\label{tab:natural}
\begin{tabular}{*1l|*1c|*1c|*1c}
\hline\noalign{\smallskip}
Method&$n$-iter : $\alpha : \beta$& $4\times$& $8\times$ \\
\noalign{\smallskip}\hline\noalign{\smallskip}
LR&-&41.7&16.6\\
Bicubic&-&41.3&11.2\\
SRGAN~\cite{ledig2016photo}&-&44.6&13.4\\
DBPN~\cite{arxiv/Haris18}&-&41.9&10.6\\
SR-FT&100$k:1:0$&52.6&22.0\\
SR-FT+&100$k:1:0$+200$k:1:0$&53.6&22.9\\
TDSR&100$k:1:0$+200$k:1:0.01$&62.2&37.5\\
\noalign{\smallskip}\hline
\end{tabular}
\end{table}

\subsection{Comparison with Different SR Methods in More Difficult Scenarios}
In realistic settings, images are afflicted by additional sources of
corruption, which can aggravate the already serious damage from
reduction in resolution. In the final set of experiments, we evaluate
ours and other methods in such settings. Here, the images (during both
train and test phases) were also degenerated by blur or noise, prior
to downscaling and processing by SR and detector. As with other
experiments, we kept the same originally pretrained SSD detector as before.

\noindent{\textbf{Blurred Images}} Every HR image was blurred by Gaussian
kernel, $\sigma=1$. In training the SR network, both in pure SR
fine-tuning and in TDSR joint optimization, the objective ($L_{rec}$)
was defined with respect to the original (clean) HR images.

The results of this experiment are shown in Table \ref{tab:blur}.
As with clean images, our proposed method outperforms all other
approaches for both scaling factors, even obtaining a small (and
likely insignificant) improvement compared to the blurry HR inputs!
This application of our method can be thought of as task-driven
deblurring by super-resolution. 

\begin{table}[t]
\caption{Analysis on blur images. Note: Original images (HR+Blur) obtained 63.3\% mAP}
\centering
\label{tab:blur}
\begin{tabular}{*1l|*1c|*1c|*1c}
\hline\noalign{\smallskip}
Method&$n$-iter : $\alpha : \beta$& $4\times$& $8\times$ \\
\noalign{\smallskip}\hline\noalign{\smallskip}
LR&-&40.1&16.2\\
Bicubic&-&42.9&11.8\\
SR-FT&-&54.7&23.9\\
SR-FT+&100$k:1:0$+200$k:1:0$&55.5&25.1\\
TDSR&100$k:1:0$+200$k:1:0.1$&63.8&39.1\\
\noalign{\smallskip}\hline
\end{tabular}
\end{table}


\noindent{\textbf{Noisy Images}} In a similar vein, we evaluate the SR
methods on images affected by Gaussian noise ($\sigma=0.1$) prior to
downscaling. Again, $L_{rec}$ penalizes error w.r.t. the clean HR
image.

The mAP on noise HR images is 57.3, an almost 20 points drop compared
to the clean HR images.
The results are shown in Table \ref{tab:noise}.
As with blur, our proposed method outperforms significantly all other
approaches for both scaling factors.
%

\begin{table}[t]
\caption{Analysis on noise images. Note: Original images (HR+Noise) obtained 57.3\% mAP}
\centering
\label{tab:noise}
\begin{tabular}{*1l|*1c|*1c|*1c}
\hline\noalign{\smallskip}
Method&$n$-iter : $\alpha : \beta$& $4\times$& $8\times$ \\
\noalign{\smallskip}\hline\noalign{\smallskip}
LR&-&39.0&14.5\\
Bicubic&-&21.2&2.84\\
SR-FT&-&41.5&11.6\\
SR-FT+&100$k:1:0$+200$k:1:0$&42.7&12.6\\
TDSR&100$k:1:0$+200$k:1:0.1$&50.1&22.7\\
\noalign{\smallskip}\hline
\end{tabular}
\end{table}


\subsection{Qualitative Analysis}\label{sec:qual}
\begin{figure}[!t]
  \begin{center}
    \begin{tabular}[c]{cccccc}
      \includegraphics[width=.15\textwidth]{images_temp/1012/1012_hr_original}
      &
        \includegraphics[width=.15\textwidth]{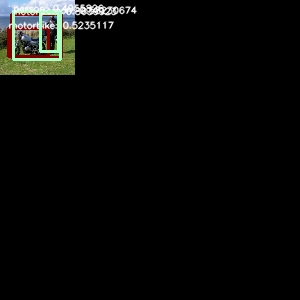}
      &
        \includegraphics[width=.15\textwidth]{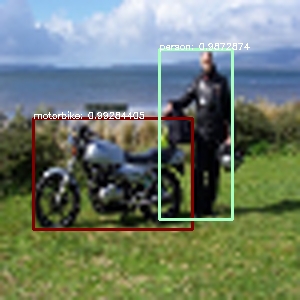}
        &
        \includegraphics[width=.15\textwidth]{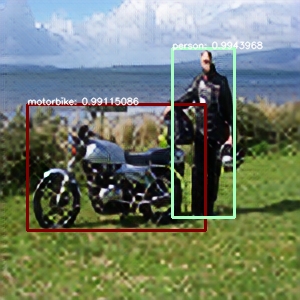}
      &
        \includegraphics[width=.15\textwidth]{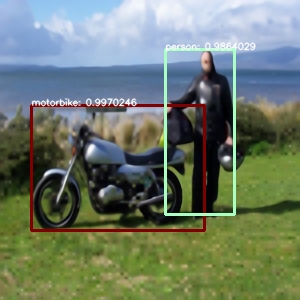}
      &
        \includegraphics[width=.15\textwidth]{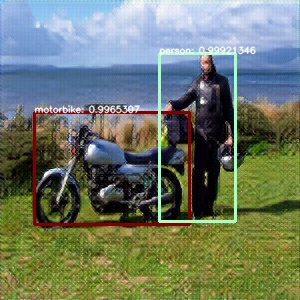}\\
        
        &
        \includegraphics[width=.15\textwidth]{images_temp/1012/1012_8x_lr_original}
      &
        \includegraphics[width=.15\textwidth]{images_temp/1012/1012_8x_bicubic_original}
        &
        \includegraphics[width=.15\textwidth]{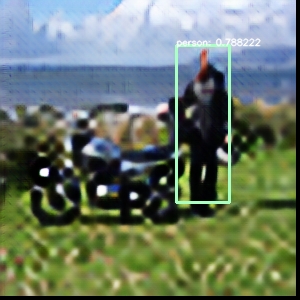}
      &
        \includegraphics[width=.15\textwidth]{images_temp/1012/1012_8x_ftplus_original}
      &
        \includegraphics[width=.15\textwidth]{images_temp/1012/1012_8x_balance_original}\\

      \includegraphics[width=.15\textwidth]{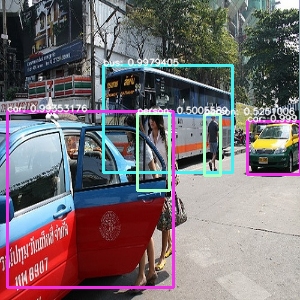}
      &
        \includegraphics[width=.15\textwidth]{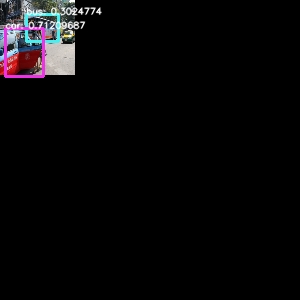}
      &
        \includegraphics[width=.15\textwidth]{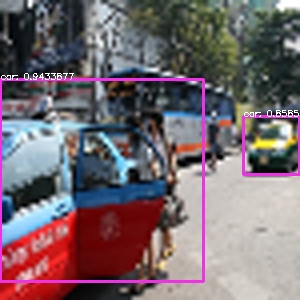}
        &
        \includegraphics[width=.15\textwidth]{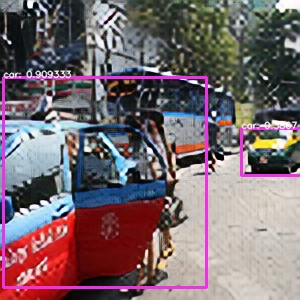}
      &
        \includegraphics[width=.15\textwidth]{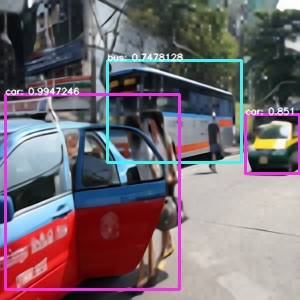}
      &
        \includegraphics[width=.15\textwidth]{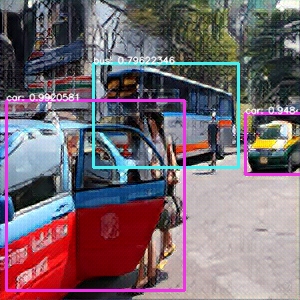}\\
        
        &
        \includegraphics[width=.15\textwidth]{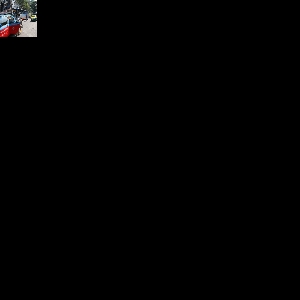}
      &
        \includegraphics[width=.15\textwidth]{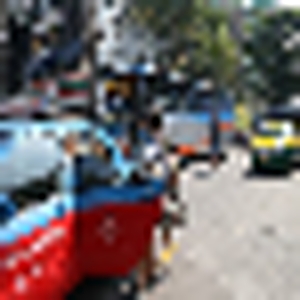}
        &
        \includegraphics[width=.15\textwidth]{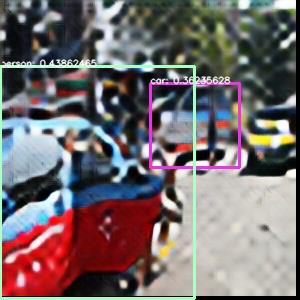}
      &
        \includegraphics[width=.15\textwidth]{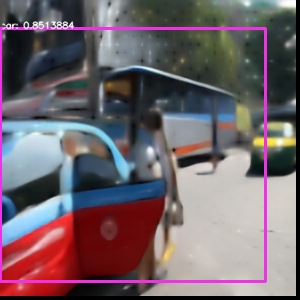}
      &
        \includegraphics[width=.15\textwidth]{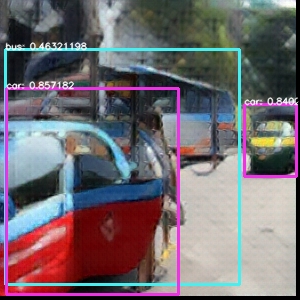}\\
        
        \includegraphics[width=.15\textwidth]{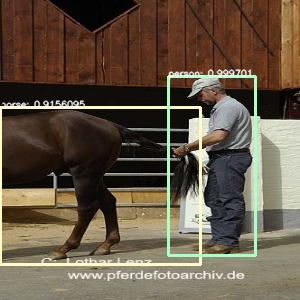}
      &
        \includegraphics[width=.15\textwidth]{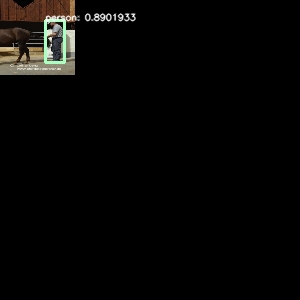}
      &
        \includegraphics[width=.15\textwidth]{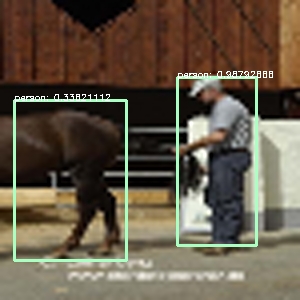}
        &
        \includegraphics[width=.15\textwidth]{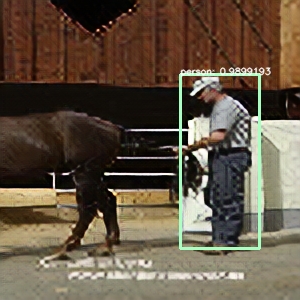}
      &
        \includegraphics[width=.15\textwidth]{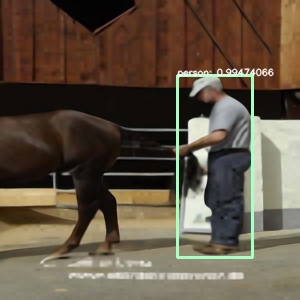}
      &
        \includegraphics[width=.15\textwidth]{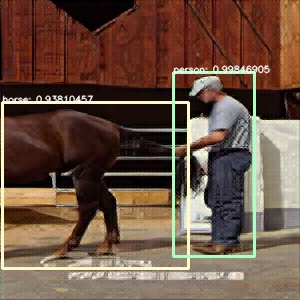}\\
        
        &
        \includegraphics[width=.15\textwidth]{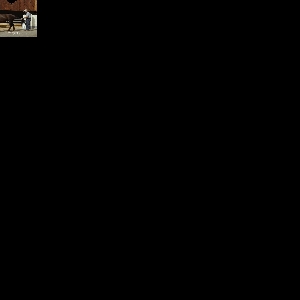}
      &
        \includegraphics[width=.15\textwidth]{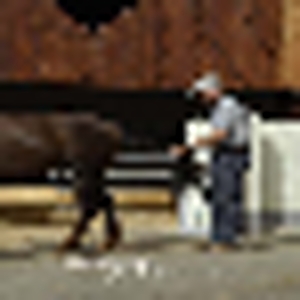}
        &
        \includegraphics[width=.15\textwidth]{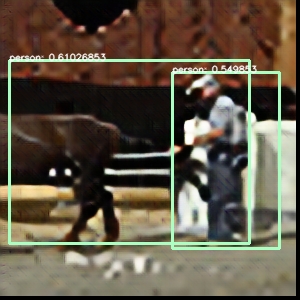}
      &
        \includegraphics[width=.15\textwidth]{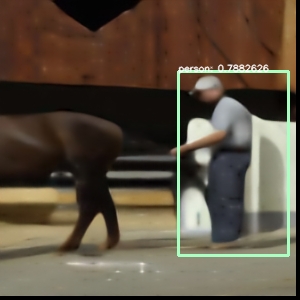}
      &
        \includegraphics[width=.15\textwidth]{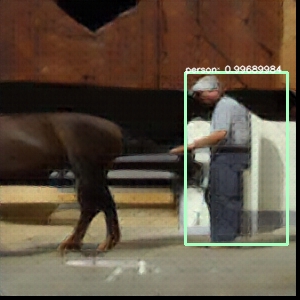}\\
        
         \includegraphics[width=.15\textwidth]{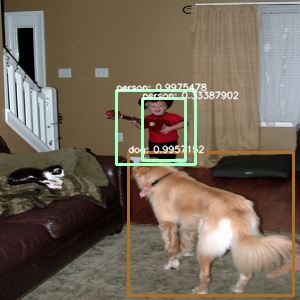}
      &
        \includegraphics[width=.15\textwidth]{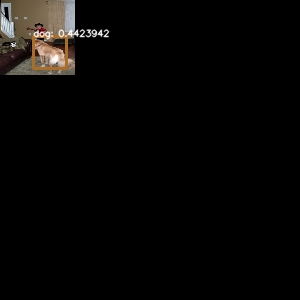}
      &
        \includegraphics[width=.15\textwidth]{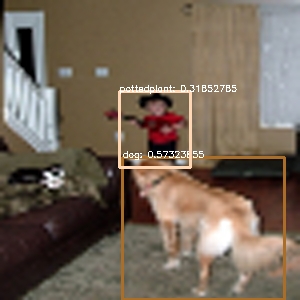}
        &
        \includegraphics[width=.15\textwidth]{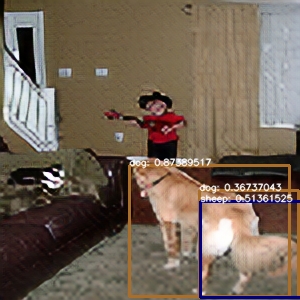}
      &
        \includegraphics[width=.15\textwidth]{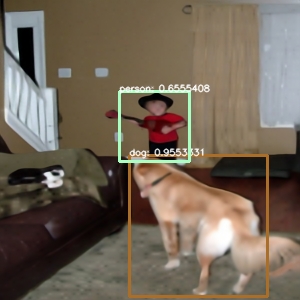}
      &
        \includegraphics[width=.15\textwidth]{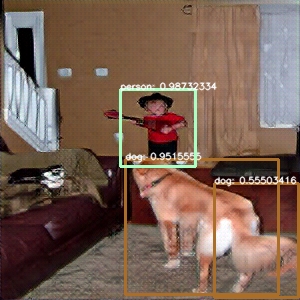}\\
        
        &
        \includegraphics[width=.15\textwidth]{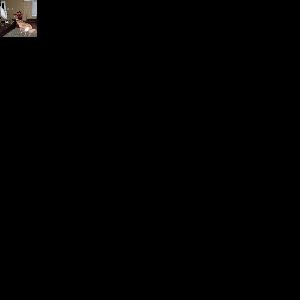}
      &
        \includegraphics[width=.15\textwidth]{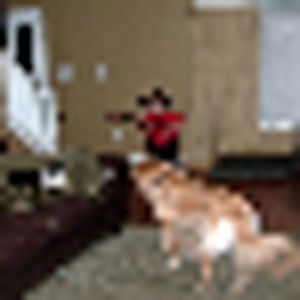}
        &
        \includegraphics[width=.15\textwidth]{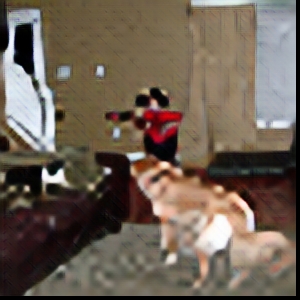}
      &
        \includegraphics[width=.15\textwidth]{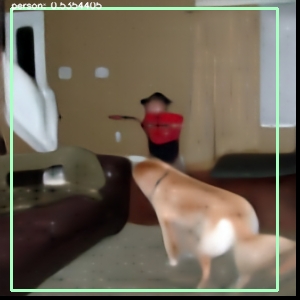}
      &
        \includegraphics[width=.15\textwidth]{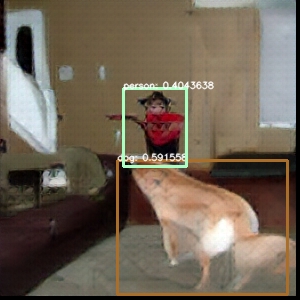}\\
      {\scriptsize (a) HR}
      &{\scriptsize (b) LR}
      &{\scriptsize (c) Bicubic}
       &{\scriptsize (d) SRGAN~\cite{ledig2016photo}}
      &{\scriptsize (e) SR-FT+}
      &{\scriptsize (f) TDSR}\\
    \end{tabular}
    \caption{Sample results for $4\times$ and $8\times$. Zoom in to
      see detection labels and scores.}
    \label{fig:natural}
  \end{center}
\end{figure}

\begin{figure}[!t]
  \begin{center}
    \begin{tabular}[c]{ccccc}
      \includegraphics[width=.18\textwidth]{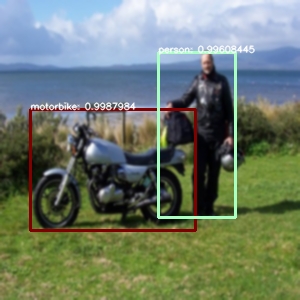}
      &
        \includegraphics[width=.18\textwidth]{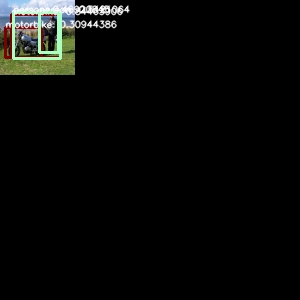}
      &
        \includegraphics[width=.18\textwidth]{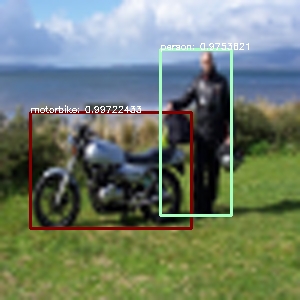}
        &
        \includegraphics[width=.18\textwidth]{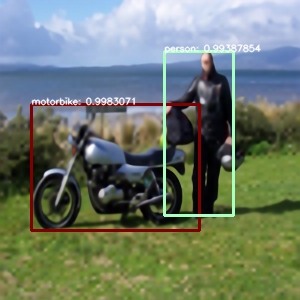}
      &
        \includegraphics[width=.18\textwidth]{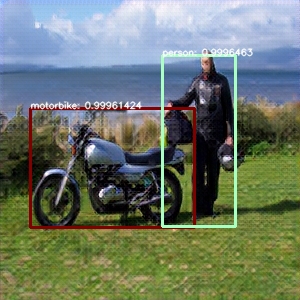}\\
        
        &
        \includegraphics[width=.18\textwidth]{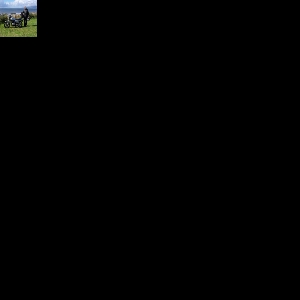}
      &
        \includegraphics[width=.18\textwidth]{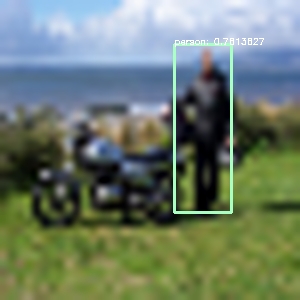}
        &
        \includegraphics[width=.18\textwidth]{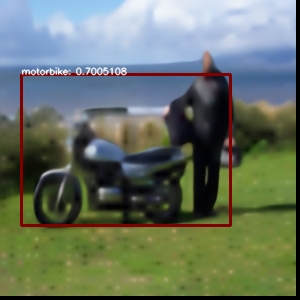}
      &
        \includegraphics[width=.18\textwidth]{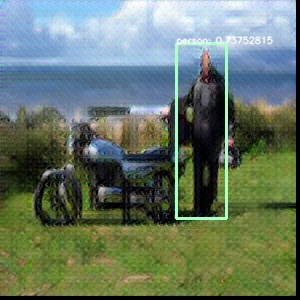}\\

      \includegraphics[width=.18\textwidth]{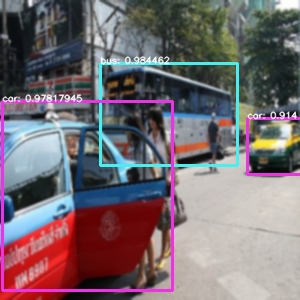}
      &
        \includegraphics[width=.18\textwidth]{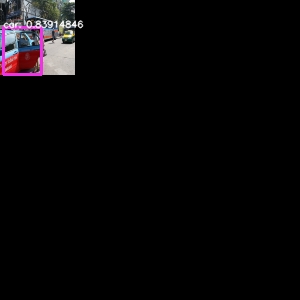}
      &
        \includegraphics[width=.18\textwidth]{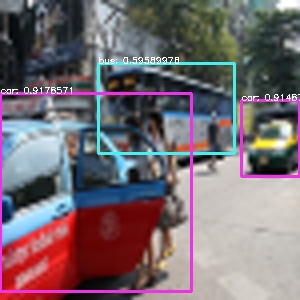}
      &
        \includegraphics[width=.18\textwidth]{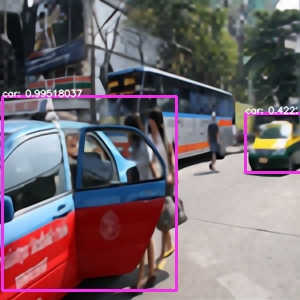}
      &
        \includegraphics[width=.18\textwidth]{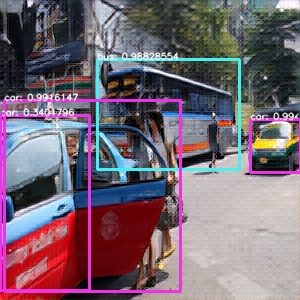}\\
        
        &
        \includegraphics[width=.18\textwidth]{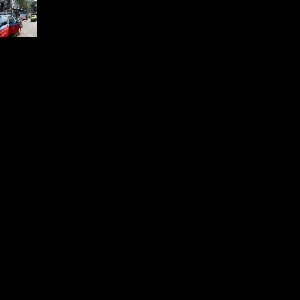}
      &
        \includegraphics[width=.18\textwidth]{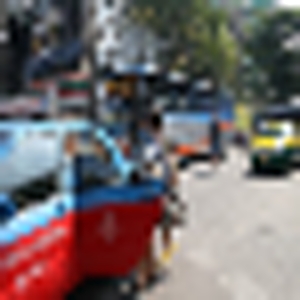}
      &
        \includegraphics[width=.18\textwidth]{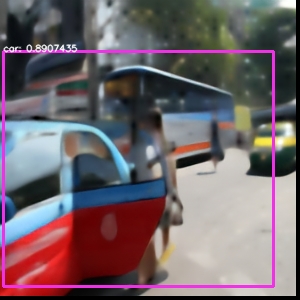}
      &
        \includegraphics[width=.18\textwidth]{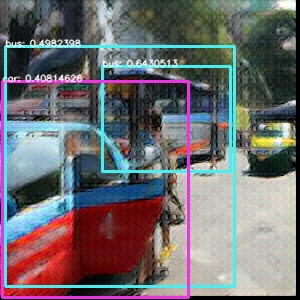}\\
        
        \includegraphics[width=.18\textwidth]{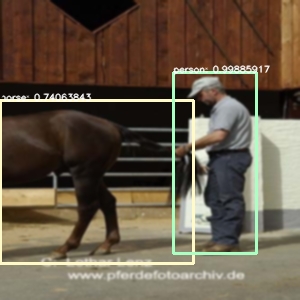}
      &
        \includegraphics[width=.18\textwidth]{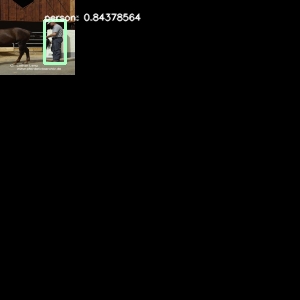}
      &
        \includegraphics[width=.18\textwidth]{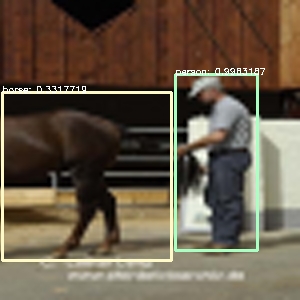}
      &
        \includegraphics[width=.18\textwidth]{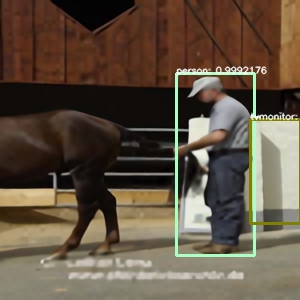}
      &
        \includegraphics[width=.18\textwidth]{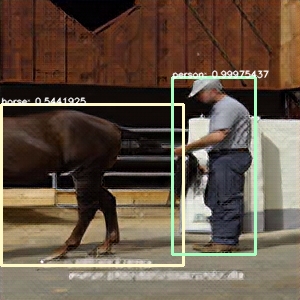}\\
        
        &
        \includegraphics[width=.18\textwidth]{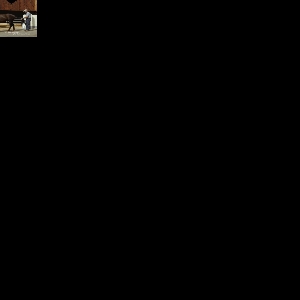}
      &
        \includegraphics[width=.18\textwidth]{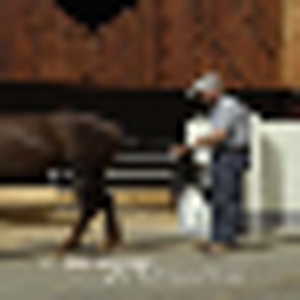}
      &
        \includegraphics[width=.18\textwidth]{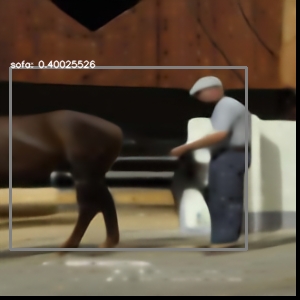}
      &
        \includegraphics[width=.18\textwidth]{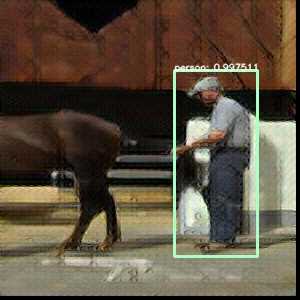}\\
        
          \includegraphics[width=.18\textwidth]{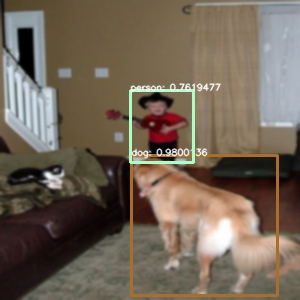}
      &
        \includegraphics[width=.18\textwidth]{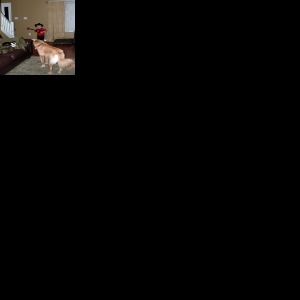}
      &
        \includegraphics[width=.18\textwidth]{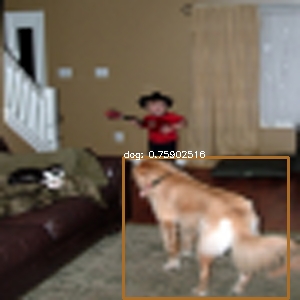}
      &
        \includegraphics[width=.18\textwidth]{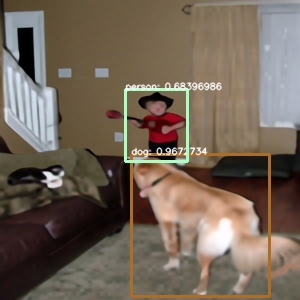}
      &
        \includegraphics[width=.18\textwidth]{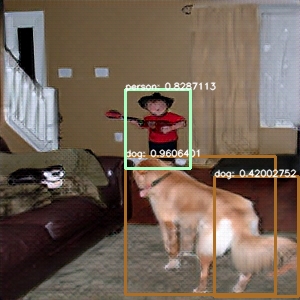}\\
        
        &
        \includegraphics[width=.18\textwidth]{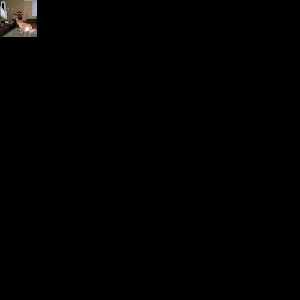}
      &
        \includegraphics[width=.18\textwidth]{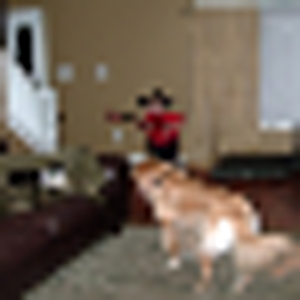}
      &
        \includegraphics[width=.18\textwidth]{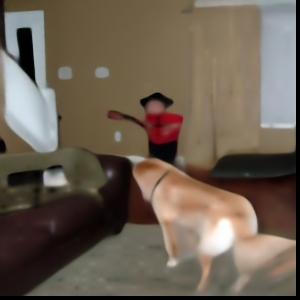}
      &
        \includegraphics[width=.18\textwidth]{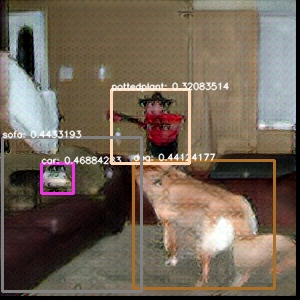}\\
      {\scriptsize (a) HR+Blur}
      &{\scriptsize (b) LR}
      &{\scriptsize (c) Bicubic}
      &{\scriptsize (d) SR-FT+}
      &{\scriptsize (e) TDSR}\\
    \end{tabular}
    \caption{Sample results on blur images for $4\times$ and
      $8\times$. Zoom in to
      see detection labels and scores.}
    \label{fig:blur}
  \end{center}
\end{figure}

\begin{figure}[!t]
  \begin{center}
    \begin{tabular}[c]{ccccc}
      \includegraphics[width=.18\textwidth]{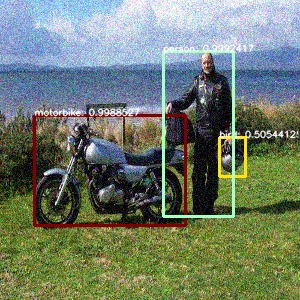}
      &
        \includegraphics[width=.18\textwidth]{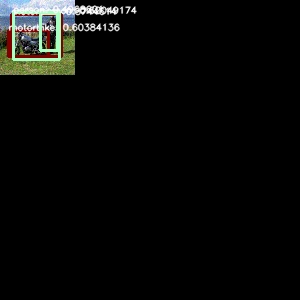}
      &
        \includegraphics[width=.18\textwidth]{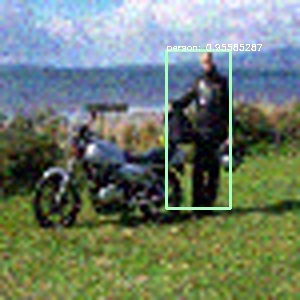}
        &
        \includegraphics[width=.18\textwidth]{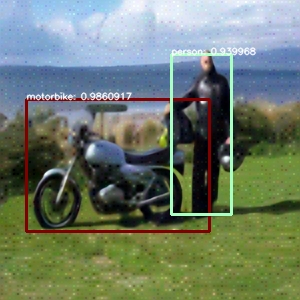}
      &
        \includegraphics[width=.18\textwidth]{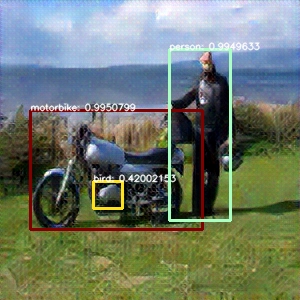}\\
        
        &
        \includegraphics[width=.18\textwidth]{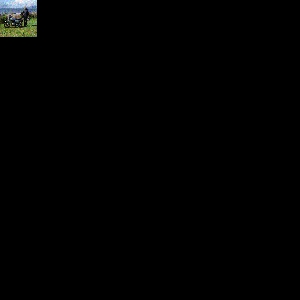}
      &
        \includegraphics[width=.18\textwidth]{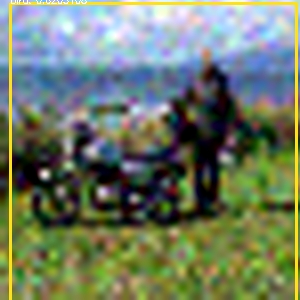}
        &
        \includegraphics[width=.18\textwidth]{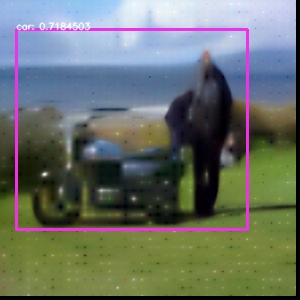}
      &
        \includegraphics[width=.18\textwidth]{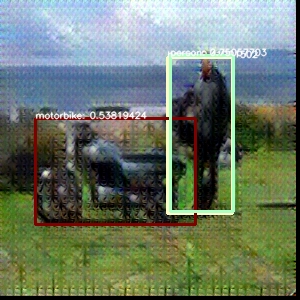}\\

      \includegraphics[width=.18\textwidth]{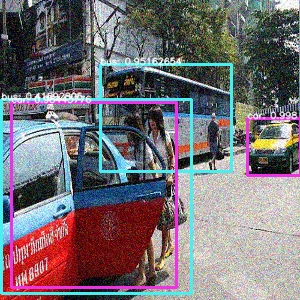}
      &
        \includegraphics[width=.18\textwidth]{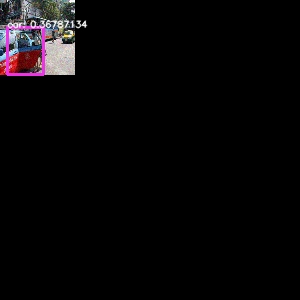}
      &
        \includegraphics[width=.18\textwidth]{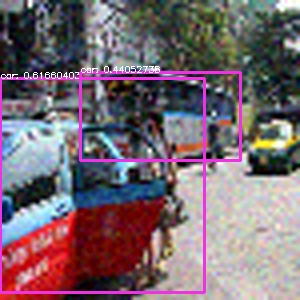}
      &
        \includegraphics[width=.18\textwidth]{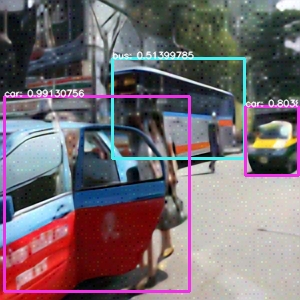}
      &
        \includegraphics[width=.18\textwidth]{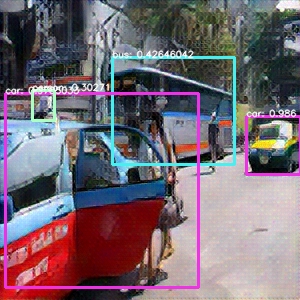}\\
        
        &
        \includegraphics[width=.18\textwidth]{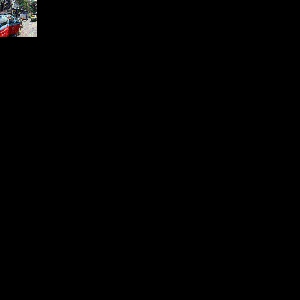}
      &
        \includegraphics[width=.18\textwidth]{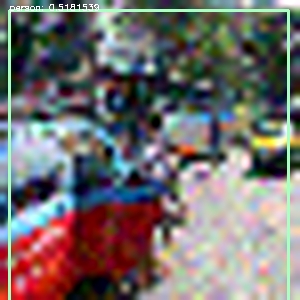}
      &
        \includegraphics[width=.18\textwidth]{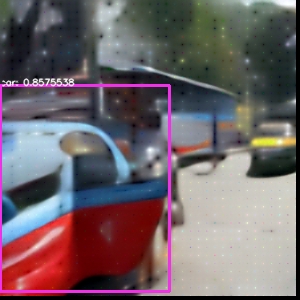}
      &
        \includegraphics[width=.18\textwidth]{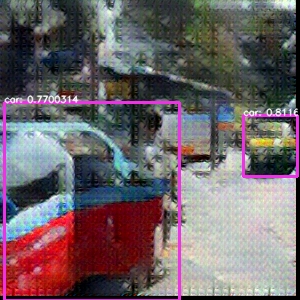}\\
        
        \includegraphics[width=.18\textwidth]{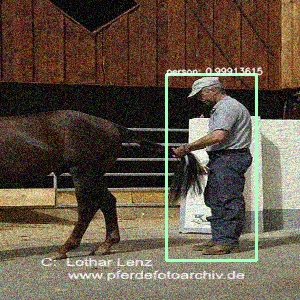}
      &
        \includegraphics[width=.18\textwidth]{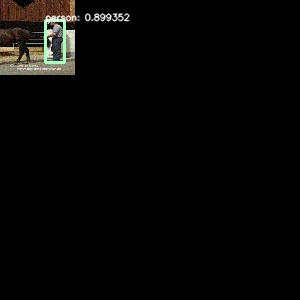}
      &
        \includegraphics[width=.18\textwidth]{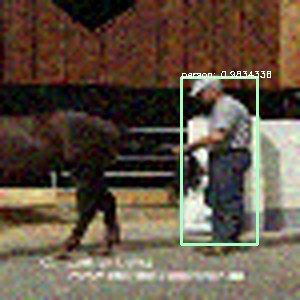}
      &
        \includegraphics[width=.18\textwidth]{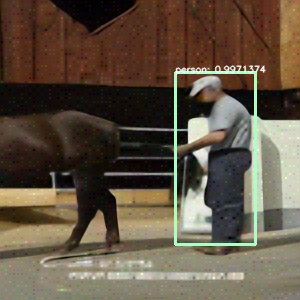}
      &
        \includegraphics[width=.18\textwidth]{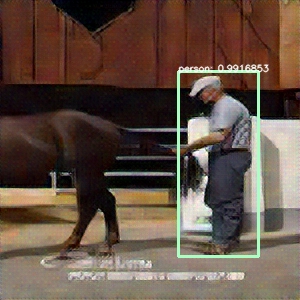}\\
        
        &
        \includegraphics[width=.18\textwidth]{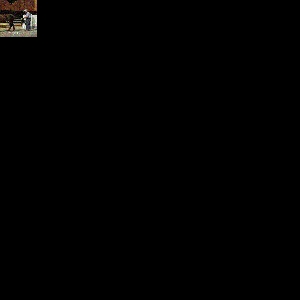}
      &
        \includegraphics[width=.18\textwidth]{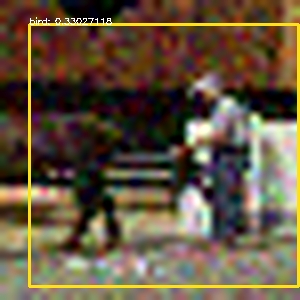}
      &
        \includegraphics[width=.18\textwidth]{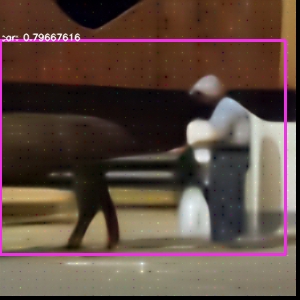}
      &
        \includegraphics[width=.18\textwidth]{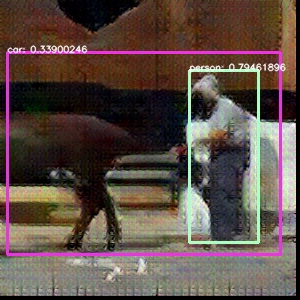}\\
        
          \includegraphics[width=.18\textwidth]{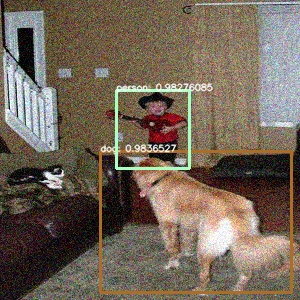}
      &
        \includegraphics[width=.18\textwidth]{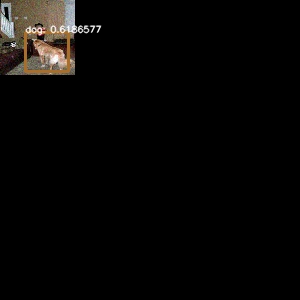}
      &
        \includegraphics[width=.18\textwidth]{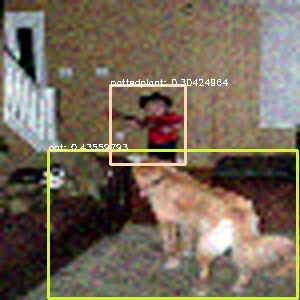}
      &
        \includegraphics[width=.18\textwidth]{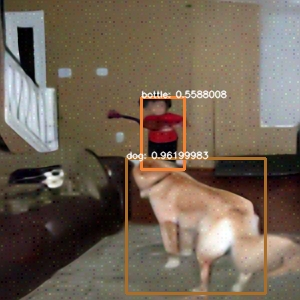}
      &
        \includegraphics[width=.18\textwidth]{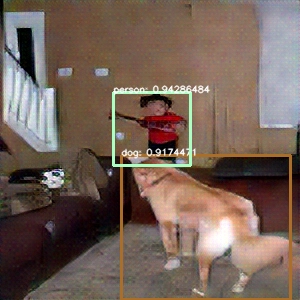}\\
        
        &
        \includegraphics[width=.18\textwidth]{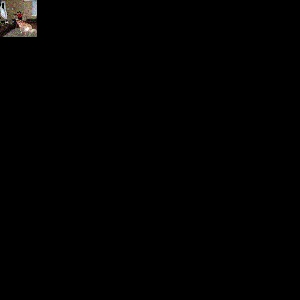}
      &
        \includegraphics[width=.18\textwidth]{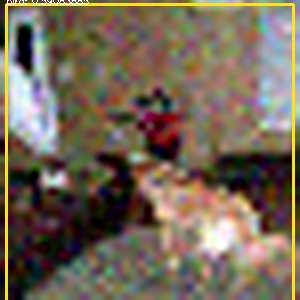}
      &
        \includegraphics[width=.18\textwidth]{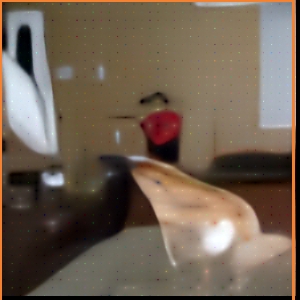}
      &
        \includegraphics[width=.18\textwidth]{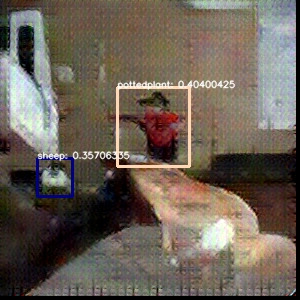}\\
      {\scriptsize (a) HR+Noise}
      &{\scriptsize (b) LR}
      &{\scriptsize (c) Bicubic}
      &{\scriptsize (d) SR-FT+}
      &{\scriptsize (e) TDSR}\\
    \end{tabular}
    \caption{Sample results on noise images for $4\times$ and
      $8\times$. Zoom in to
      see detection labels and scores.}
    \label{fig:noise}
  \end{center}
\end{figure}

Figures~\ref{fig:natural},~\ref{fig:blur},~and~\ref{fig:noise} show
examples of our results compared with those of other methods.
The results for SRGAN~\cite{ledig2016photo} and SR-FT+ sometimes confuse
the detector and recognize it as different object classes, again
indicating that optimizing $L_{rec}$ and high PSNR do not necessarily
correlate with the accuracy. Meanwhile, unique pattern that produced
by our proposed optimization helps the detector to recognize the
objects better. Note that the TDSR does produce, in many images,
 artifacts somewhat reminiscent of those in
DeepDream~\cite{deepdream}, but those are mild, and are offset by
a drastically increased detection accuracy.

\section{Conclusions}\label{sec:conclusion}

We have proposed a novel objective for training super-resolution: a
compound loss that caters to the downstream semantic task, and not
just to the pixel-wise image reconstruction task as traditionally
done.
Our results, which consistently exceed alternative SR methods in all
conditions, indicate that modern end-to-end training enables joint
optimization of tasks what has traditionally been separated into
low-level vision (super-resolution) and high-level vision (object
detection). These results also suggest some avenues for future
work. The first is to investigate task-driven SR methods for
additional visual tasks, such as semantic segmentation, image
captioning, etc. A complementary direction is to extend the
task-driven formulation to other image reconstruction and enhancement
tools. For instance, we have demonstrated some success in ``deblurring by SR'', and
one can expect further improvement when using a properly designed
deblurring network combined with task-driven objectives. Finally, the
community may be well served by a continuing quest for better image
quality metrics, to replace or augment simplistic reconstruction
losses such as PSNR; in this context we believe adversarial loss
functions to be promising.

\clearpage
\appendix
\noindent{\large{APPENDIX: Supplementary Material}}
\section{Networks Architecture}
\begin{figure}[!h]
\centering
\includegraphics[width=12cm]{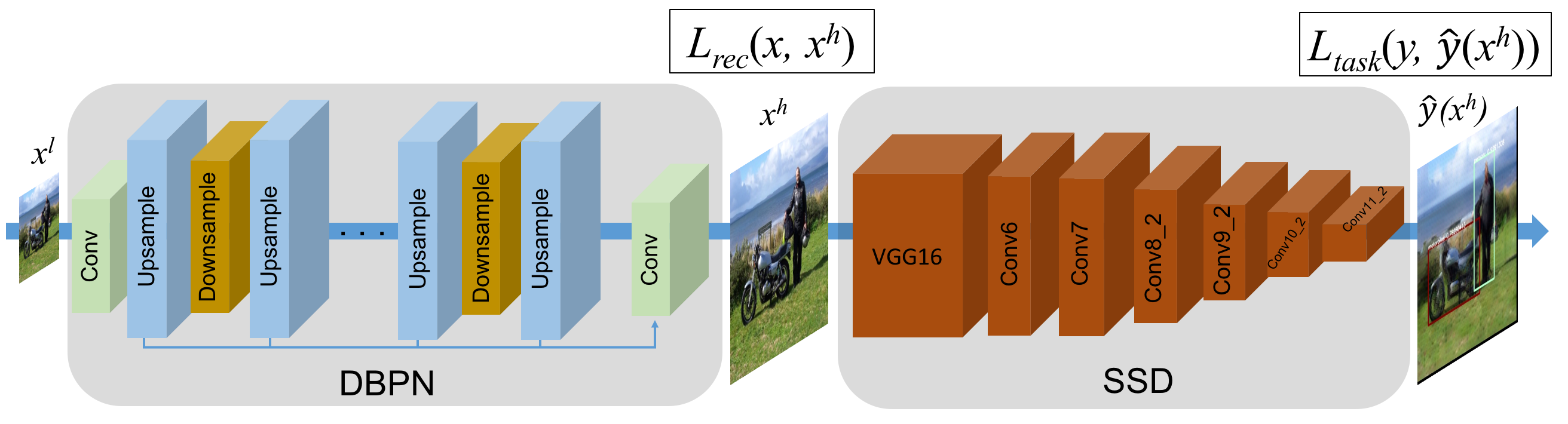}
\caption{Network Architecture}
\label{figure:architecture}
\end{figure}

Our method relies on two sequential building blocks: Super-resolution (DBPN~\cite{arxiv/Haris18}) and Task Network (SSD~\cite{DBLP:conf/eccv/LiuAESRFB16}).
All network configuration remains the same as the proposal from the original author.
On Fig.~\ref{figure:architecture}, the SR network transforms a low-resolution image $x^l$ to a high-resolution
image $x^h$. Then, the task network takes an image $x$ from SR network to produce prediction $\widehat{y}(x^h)$.

\section{Graphs on mAP and PSNR}
\begin{figure}[!t]
\centering
\includegraphics[width=12cm]{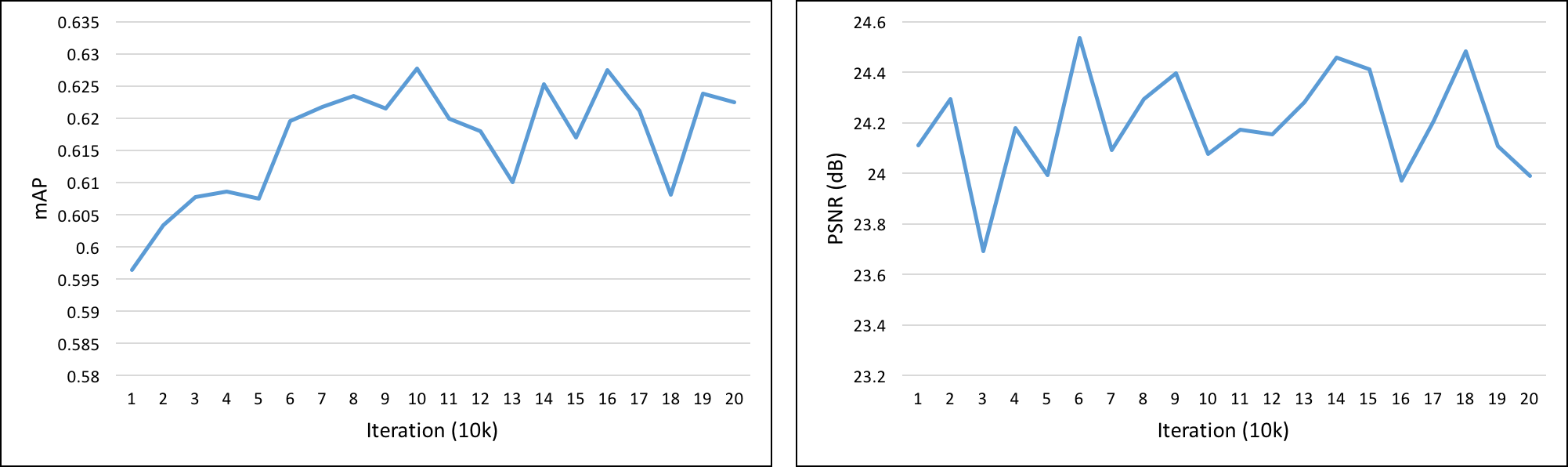}
\caption{Graph of mAP and PSNR on $4\times$}
\label{figure:map_psnr_4x}
\end{figure}

\begin{figure}[!t]
\centering
\includegraphics[width=12cm]{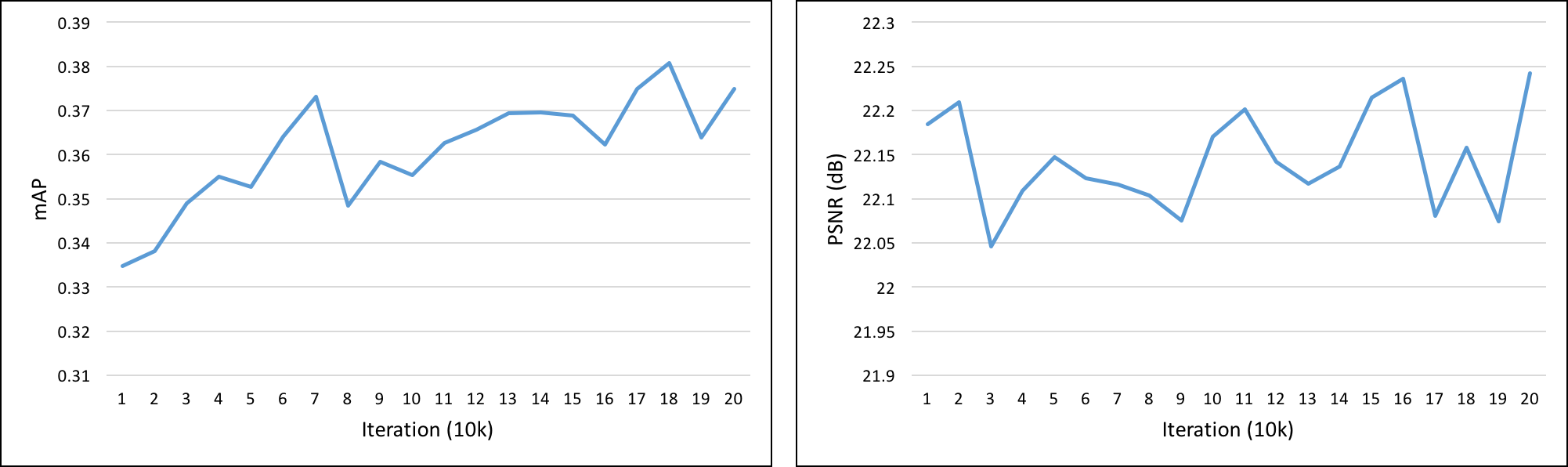}
\caption{Graph of mAP and PSNR on $8\times$}
\label{figure:map_psnr_8x}
\end{figure}

Figure~\ref{figure:map_psnr_4x}~and~\ref{figure:map_psnr_8x} show graphs where the vertical and horizontal axes denote mAP/PSNR and iterations, respectively, on $4\times$ and $8\times$ in balance setting. It shows that the balance setting successfully increases the accuracy (mAP) while maintaining a good quality of images (PSNR).

\section{Visual Results}
\begin{figure}[!t]
  \begin{center}
    \begin{tabular}[c]{cccccc}
      \includegraphics[width=.15\textwidth]{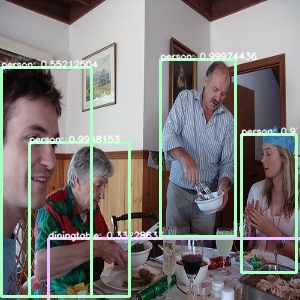}
      &
        \includegraphics[width=.15\textwidth]{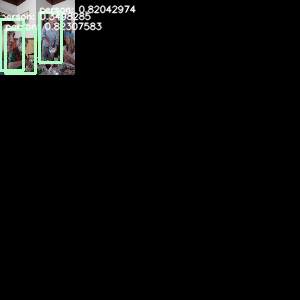}
      &
        \includegraphics[width=.15\textwidth]{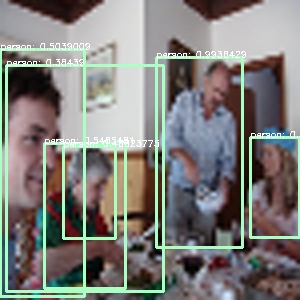}
        &
        \includegraphics[width=.15\textwidth]{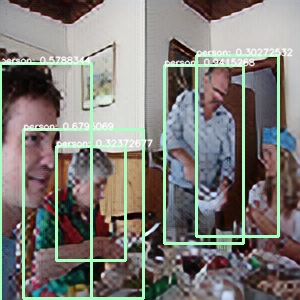}
      &
        \includegraphics[width=.15\textwidth]{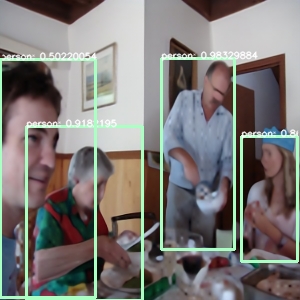}
      &
        \includegraphics[width=.15\textwidth]{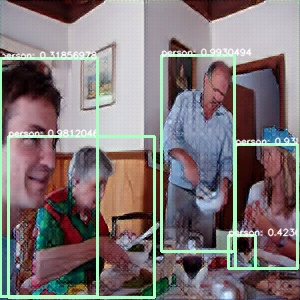}\\
        
        &
        \includegraphics[width=.15\textwidth]{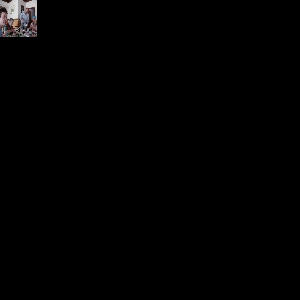}
      &
        \includegraphics[width=.15\textwidth]{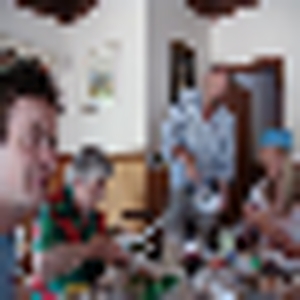}
        &
        \includegraphics[width=.15\textwidth]{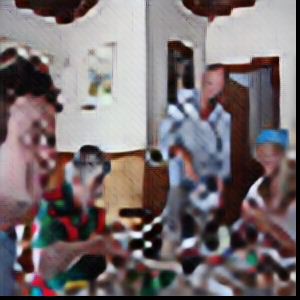}
      &
        \includegraphics[width=.15\textwidth]{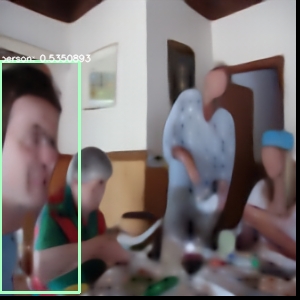}
      &
        \includegraphics[width=.15\textwidth]{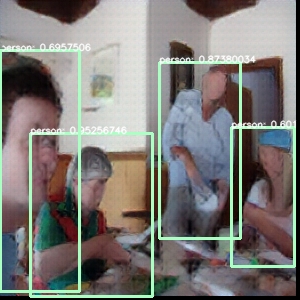}\\

      \includegraphics[width=.15\textwidth]{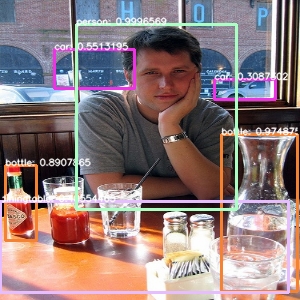}
      &
        \includegraphics[width=.15\textwidth]{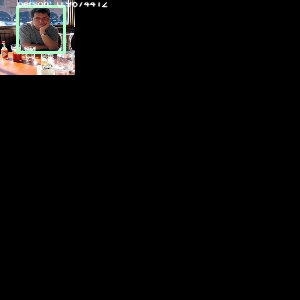}
      &
        \includegraphics[width=.15\textwidth]{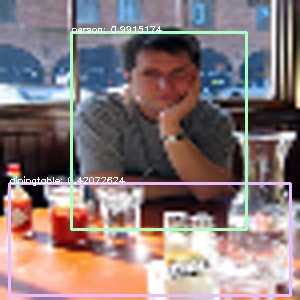}
        &
        \includegraphics[width=.15\textwidth]{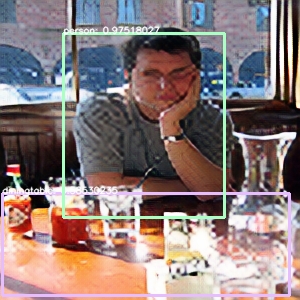}
      &
        \includegraphics[width=.15\textwidth]{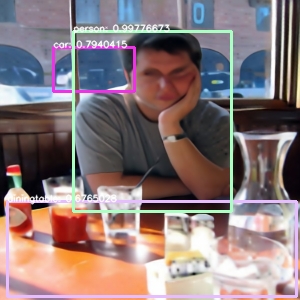}
      &
        \includegraphics[width=.15\textwidth]{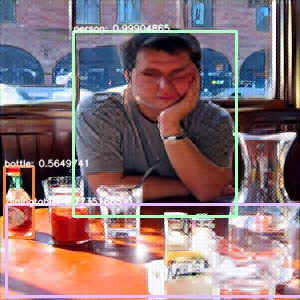}\\
        
        &
        \includegraphics[width=.15\textwidth]{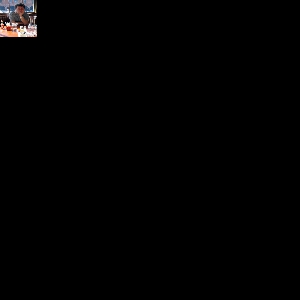}
      &
        \includegraphics[width=.15\textwidth]{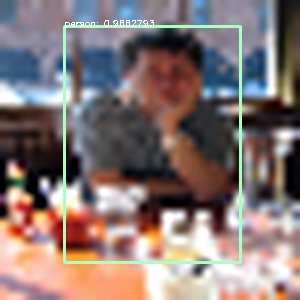}
        &
        \includegraphics[width=.15\textwidth]{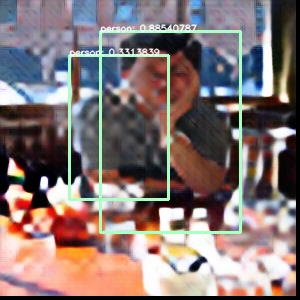}
      &
        \includegraphics[width=.15\textwidth]{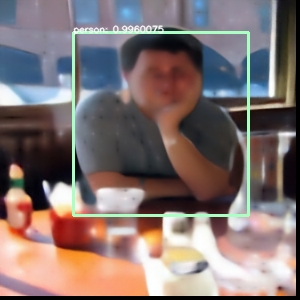}
      &
        \includegraphics[width=.15\textwidth]{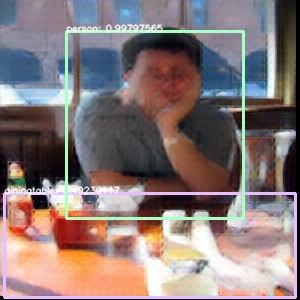}\\
        
        \includegraphics[width=.15\textwidth]{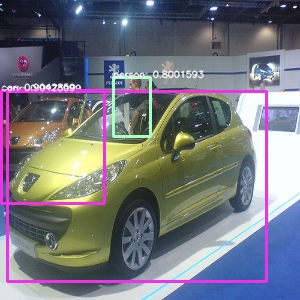}
      &
        \includegraphics[width=.15\textwidth]{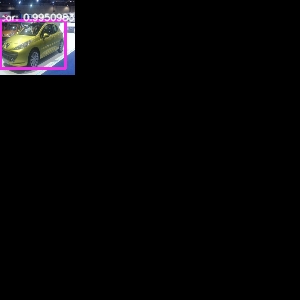}
      &
        \includegraphics[width=.15\textwidth]{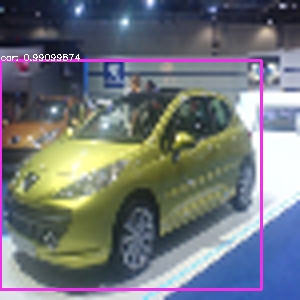}
        &
        \includegraphics[width=.15\textwidth]{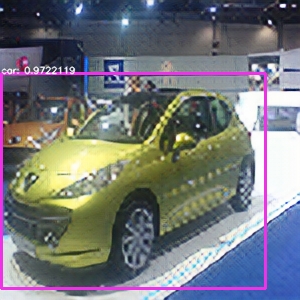}
      &
        \includegraphics[width=.15\textwidth]{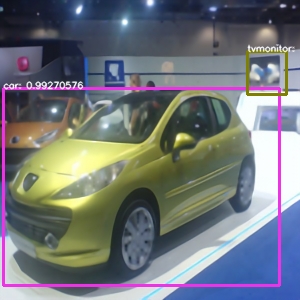}
      &
        \includegraphics[width=.15\textwidth]{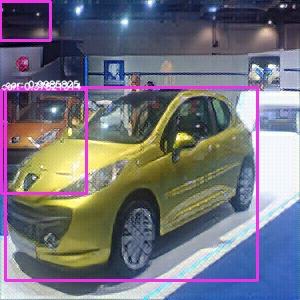}\\
        
        &
        \includegraphics[width=.15\textwidth]{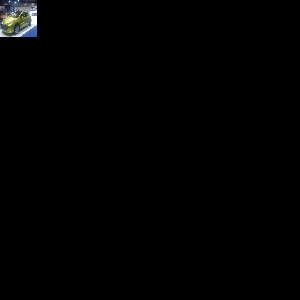}
      &
        \includegraphics[width=.15\textwidth]{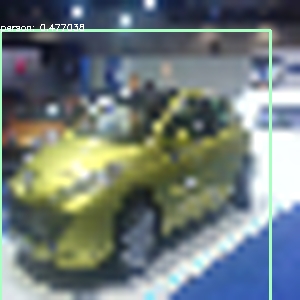}
        &
        \includegraphics[width=.15\textwidth]{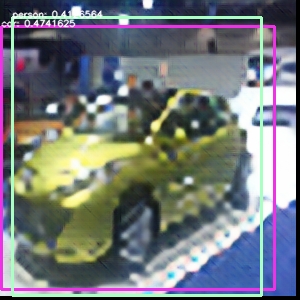}
      &
        \includegraphics[width=.15\textwidth]{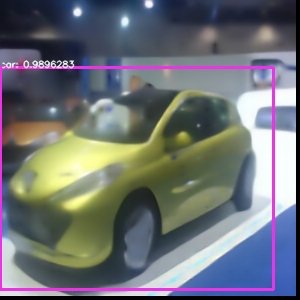}
      &
        \includegraphics[width=.15\textwidth]{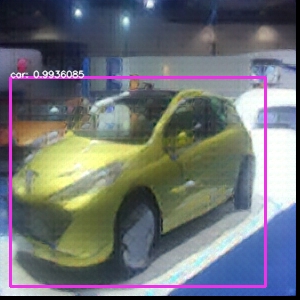}\\
        
         \includegraphics[width=.15\textwidth]{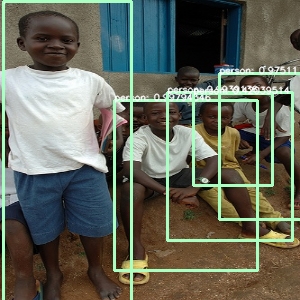}
      &
        \includegraphics[width=.15\textwidth]{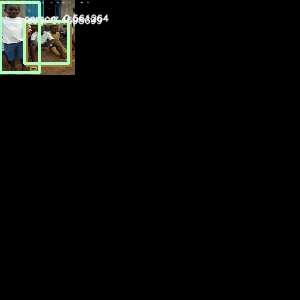}
      &
        \includegraphics[width=.15\textwidth]{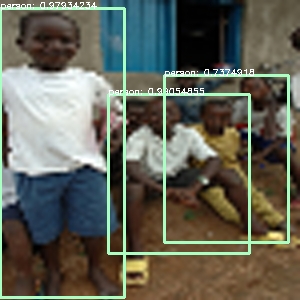}
        &
        \includegraphics[width=.15\textwidth]{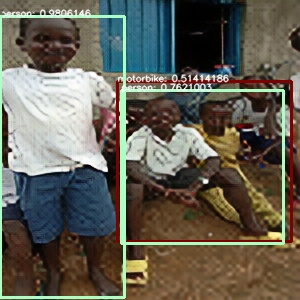}
      &
        \includegraphics[width=.15\textwidth]{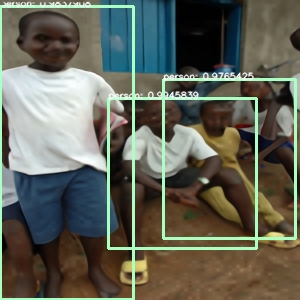}
      &
        \includegraphics[width=.15\textwidth]{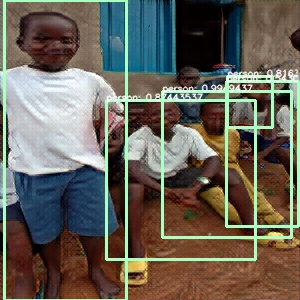}\\
        
        &
        \includegraphics[width=.15\textwidth]{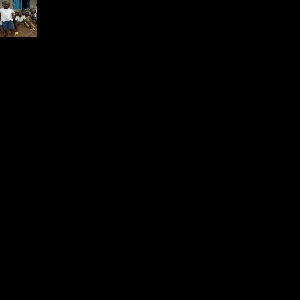}
      &
        \includegraphics[width=.15\textwidth]{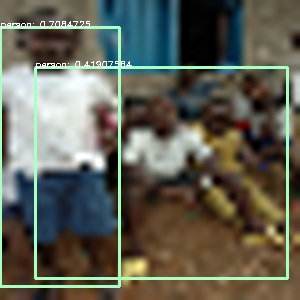}
        &
        \includegraphics[width=.15\textwidth]{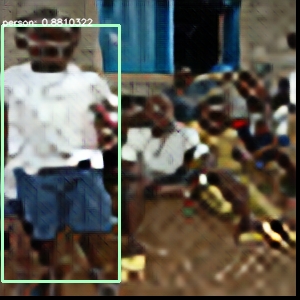}
      &
        \includegraphics[width=.15\textwidth]{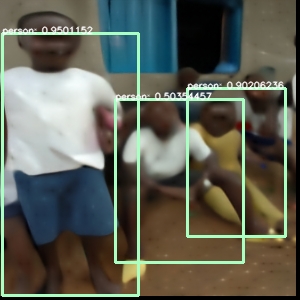}
      &
        \includegraphics[width=.15\textwidth]{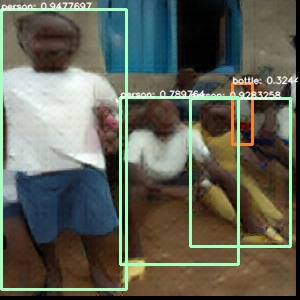}\\
      {\scriptsize (a) HR}
      &{\scriptsize (b) LR}
      &{\scriptsize (c) Bicubic}
       &{\scriptsize (d) SRGAN~\cite{ledig2016photo}}
      &{\scriptsize (e) SR-FT+}
      &{\scriptsize (f) TDSR}\\
    \end{tabular}
    \caption{Sample results for $4\times$ and $8\times$. Zoom in to
      see detection labels and scores.}
    \label{fig:natural1}
  \end{center}
\end{figure}

\begin{figure}[!t]
  \begin{center}
    \begin{tabular}[c]{cccccc}
      \includegraphics[width=.15\textwidth]{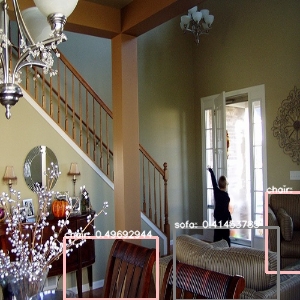}
      &
        \includegraphics[width=.15\textwidth]{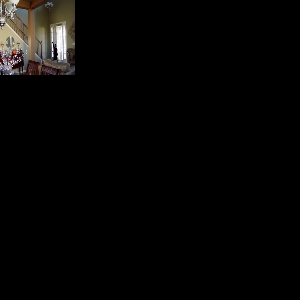}
      &
        \includegraphics[width=.15\textwidth]{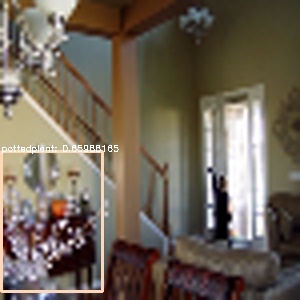}
        &
        \includegraphics[width=.15\textwidth]{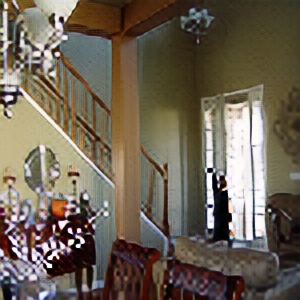}
      &
        \includegraphics[width=.15\textwidth]{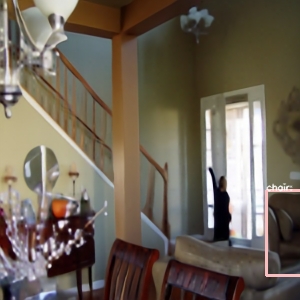}
      &
        \includegraphics[width=.15\textwidth]{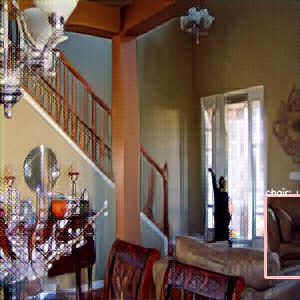}\\
        
        &
        \includegraphics[width=.15\textwidth]{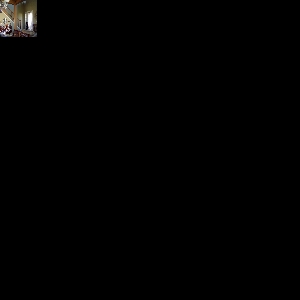}
      &
        \includegraphics[width=.15\textwidth]{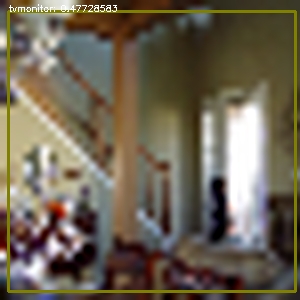}
        &
        \includegraphics[width=.15\textwidth]{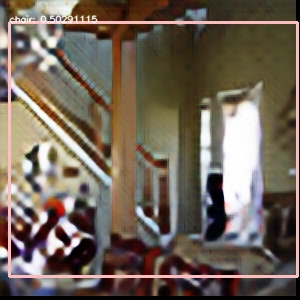}
      &
        \includegraphics[width=.15\textwidth]{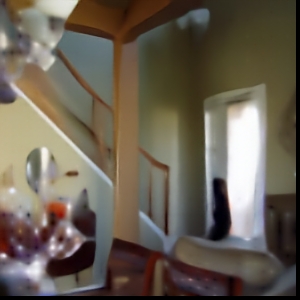}
      &
        \includegraphics[width=.15\textwidth]{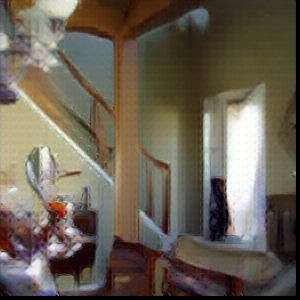}\\

      \includegraphics[width=.15\textwidth]{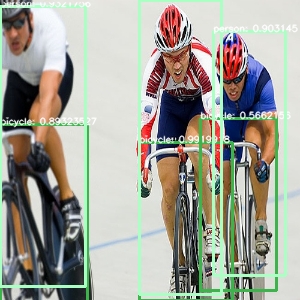}
      &
        \includegraphics[width=.15\textwidth]{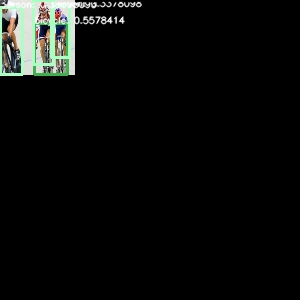}
      &
        \includegraphics[width=.15\textwidth]{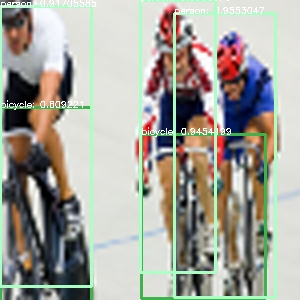}
        &
        \includegraphics[width=.15\textwidth]{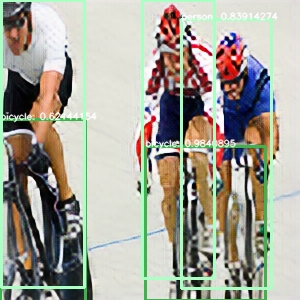}
      &
        \includegraphics[width=.15\textwidth]{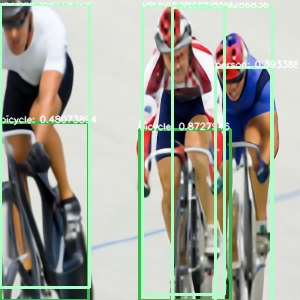}
      &
        \includegraphics[width=.15\textwidth]{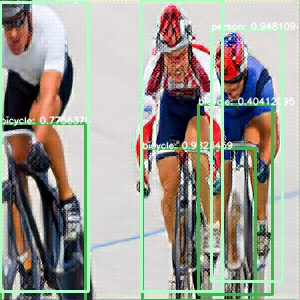}\\
        
        &
        \includegraphics[width=.15\textwidth]{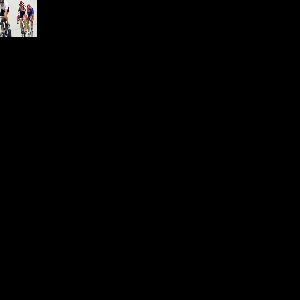}
      &
        \includegraphics[width=.15\textwidth]{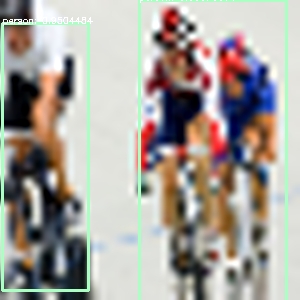}
        &
        \includegraphics[width=.15\textwidth]{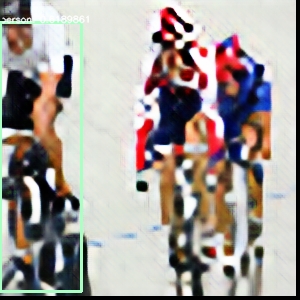}
      &
        \includegraphics[width=.15\textwidth]{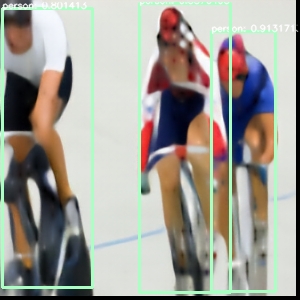}
      &
        \includegraphics[width=.15\textwidth]{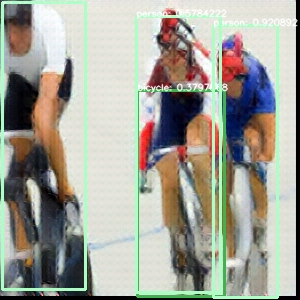}\\
        
        \includegraphics[width=.15\textwidth]{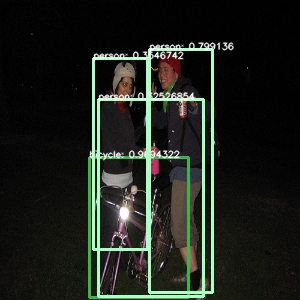}
      &
        \includegraphics[width=.15\textwidth]{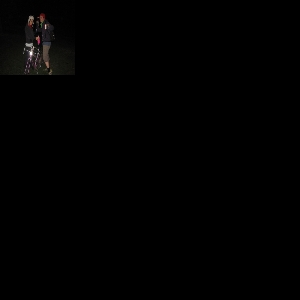}
      &
        \includegraphics[width=.15\textwidth]{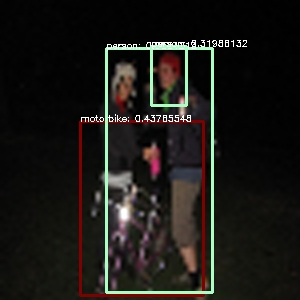}
        &
        \includegraphics[width=.15\textwidth]{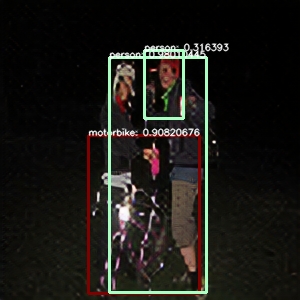}
      &
        \includegraphics[width=.15\textwidth]{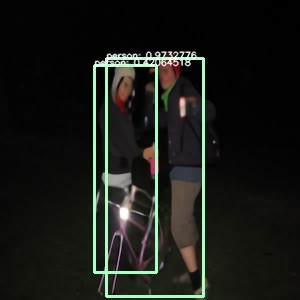}
      &
        \includegraphics[width=.15\textwidth]{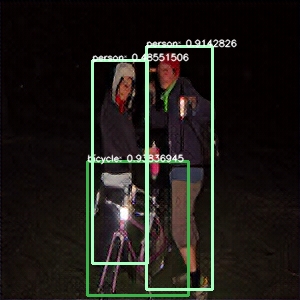}\\
        
        &
        \includegraphics[width=.15\textwidth]{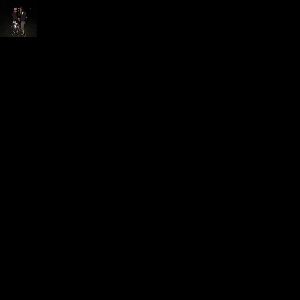}
      &
        \includegraphics[width=.15\textwidth]{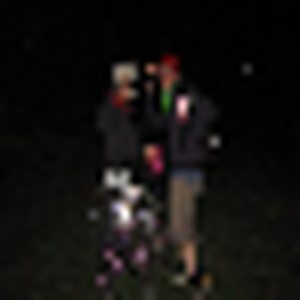}
        &
        \includegraphics[width=.15\textwidth]{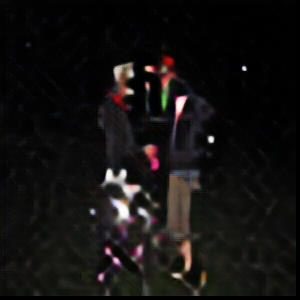}
      &
        \includegraphics[width=.15\textwidth]{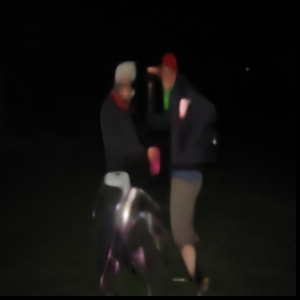}
      &
        \includegraphics[width=.15\textwidth]{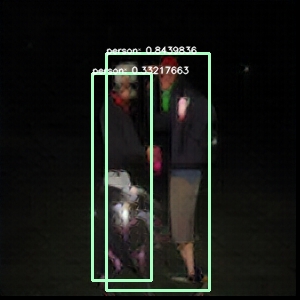}\\
        
         \includegraphics[width=.15\textwidth]{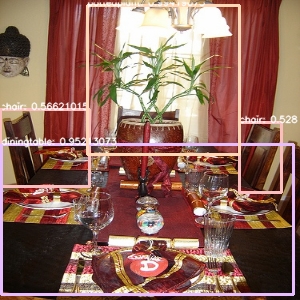}
      &
        \includegraphics[width=.15\textwidth]{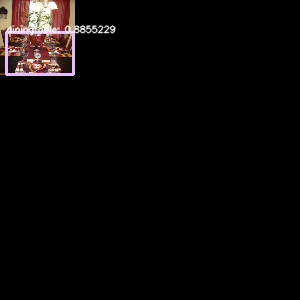}
      &
        \includegraphics[width=.15\textwidth]{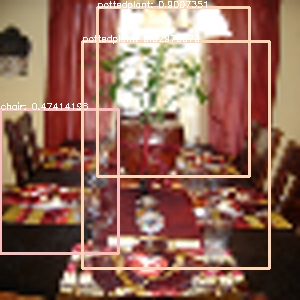}
        &
        \includegraphics[width=.15\textwidth]{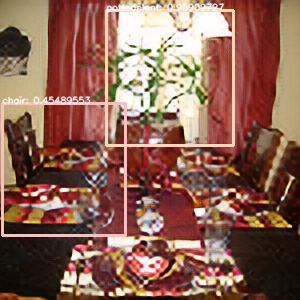}
      &
        \includegraphics[width=.15\textwidth]{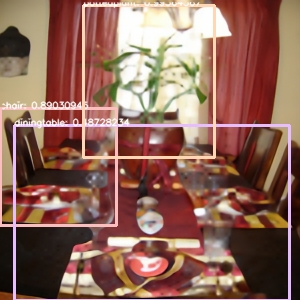}
      &
        \includegraphics[width=.15\textwidth]{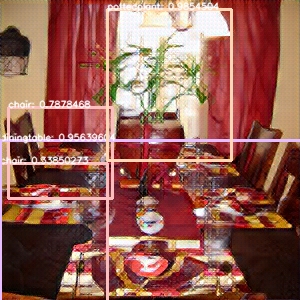}\\
        
        &
        \includegraphics[width=.15\textwidth]{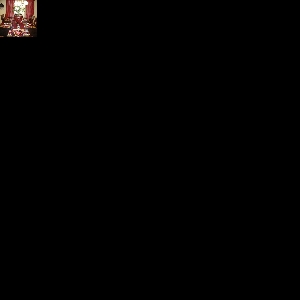}
      &
        \includegraphics[width=.15\textwidth]{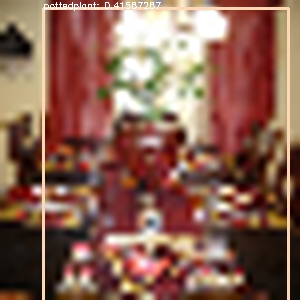}
        &
        \includegraphics[width=.15\textwidth]{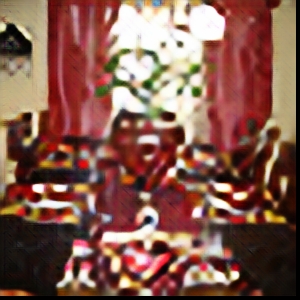}
      &
        \includegraphics[width=.15\textwidth]{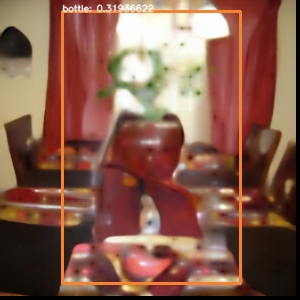}
      &
        \includegraphics[width=.15\textwidth]{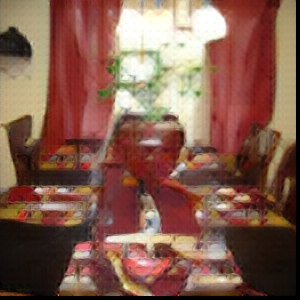}\\
      {\scriptsize (a) HR}
      &{\scriptsize (b) LR}
      &{\scriptsize (c) Bicubic}
       &{\scriptsize (d) SRGAN~\cite{ledig2016photo}}
      &{\scriptsize (e) SR-FT+}
      &{\scriptsize (f) TDSR}\\
    \end{tabular}
    \caption{Sample results for $4\times$ and $8\times$. Zoom in to
      see detection labels and scores.}
    \label{fig:natural2}
  \end{center}
\end{figure}

\begin{figure}[!t]
  \begin{center}
    \begin{tabular}[c]{ccccc}
      \includegraphics[width=.18\textwidth]{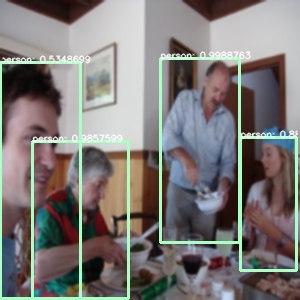}
      &
        \includegraphics[width=.18\textwidth]{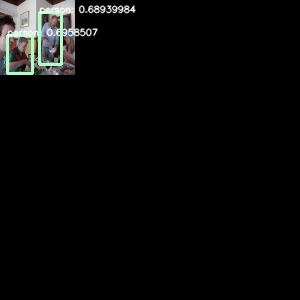}
      &
        \includegraphics[width=.18\textwidth]{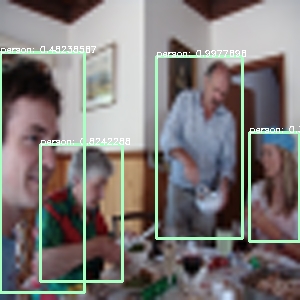}
        &
        \includegraphics[width=.18\textwidth]{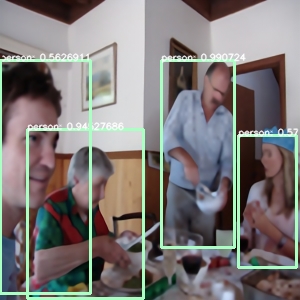}
      &
        \includegraphics[width=.18\textwidth]{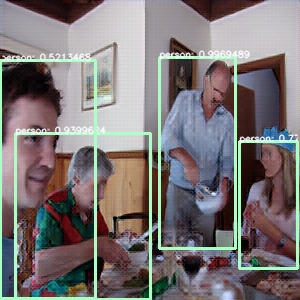}\\
        
        &
        \includegraphics[width=.18\textwidth]{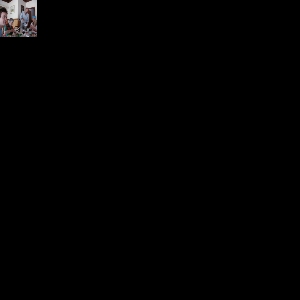}
      &
        \includegraphics[width=.18\textwidth]{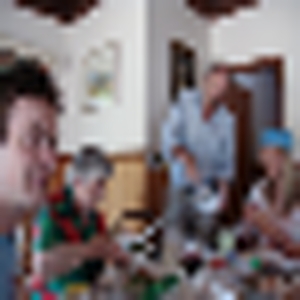}
        &
        \includegraphics[width=.18\textwidth]{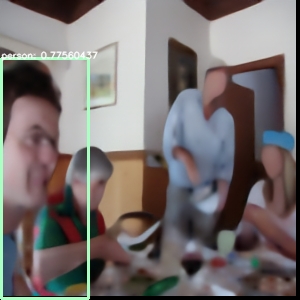}
      &
        \includegraphics[width=.18\textwidth]{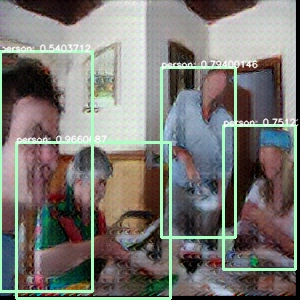}\\

      \includegraphics[width=.18\textwidth]{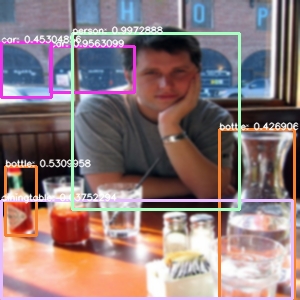}
      &
        \includegraphics[width=.18\textwidth]{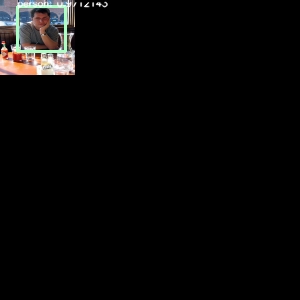}
      &
        \includegraphics[width=.18\textwidth]{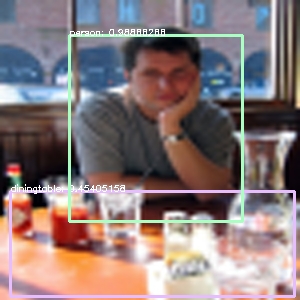}
      &
        \includegraphics[width=.18\textwidth]{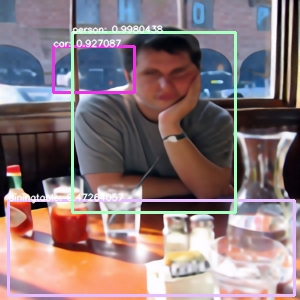}
      &
        \includegraphics[width=.18\textwidth]{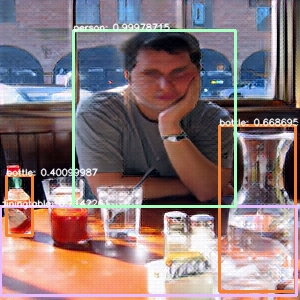}\\
        
        &
        \includegraphics[width=.18\textwidth]{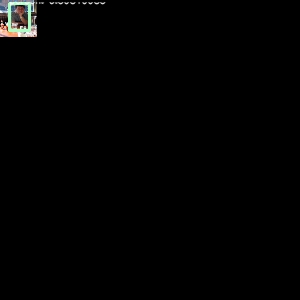}
      &
        \includegraphics[width=.18\textwidth]{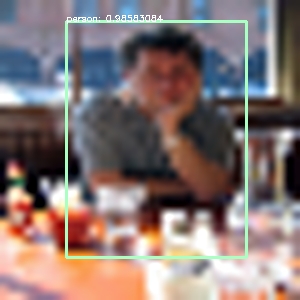}
      &
        \includegraphics[width=.18\textwidth]{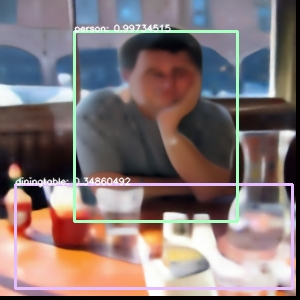}
      &
        \includegraphics[width=.18\textwidth]{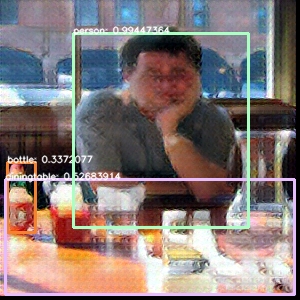}\\
        
        \includegraphics[width=.18\textwidth]{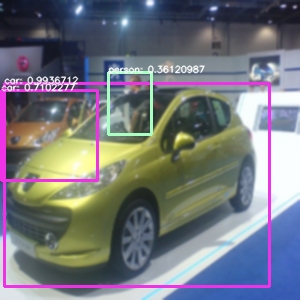}
      &
        \includegraphics[width=.18\textwidth]{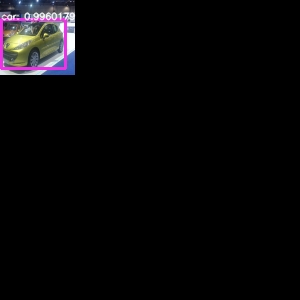}
      &
        \includegraphics[width=.18\textwidth]{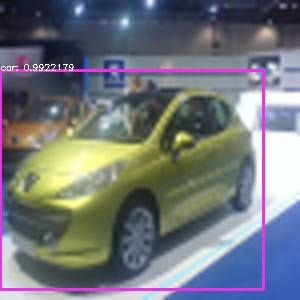}
      &
        \includegraphics[width=.18\textwidth]{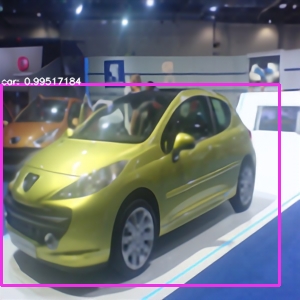}
      &
        \includegraphics[width=.18\textwidth]{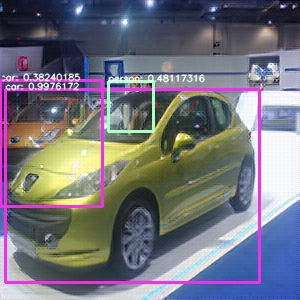}\\
        
        &
        \includegraphics[width=.18\textwidth]{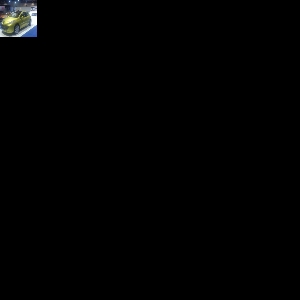}
      &
        \includegraphics[width=.18\textwidth]{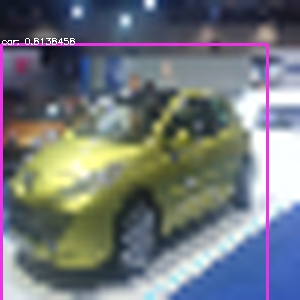}
      &
        \includegraphics[width=.18\textwidth]{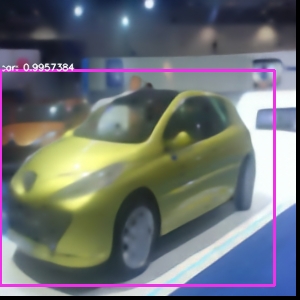}
      &
        \includegraphics[width=.18\textwidth]{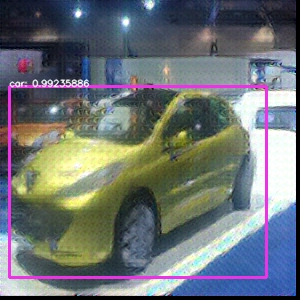}\\
        
          \includegraphics[width=.18\textwidth]{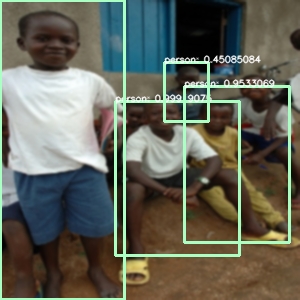}
      &
        \includegraphics[width=.18\textwidth]{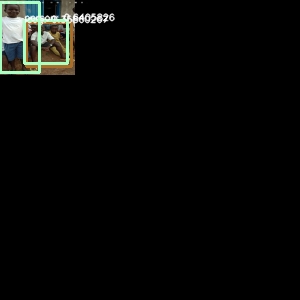}
      &
        \includegraphics[width=.18\textwidth]{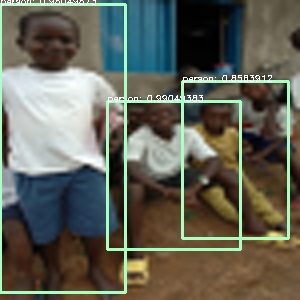}
      &
        \includegraphics[width=.18\textwidth]{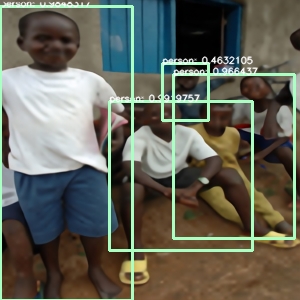}
      &
        \includegraphics[width=.18\textwidth]{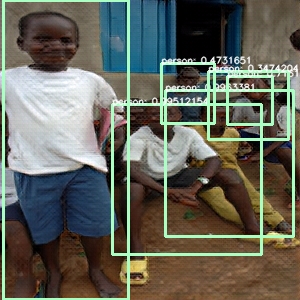}\\
        
        &
        \includegraphics[width=.18\textwidth]{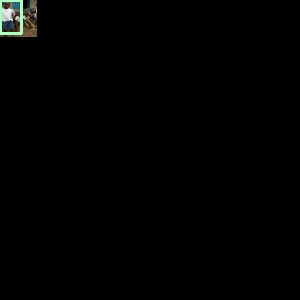}
      &
        \includegraphics[width=.18\textwidth]{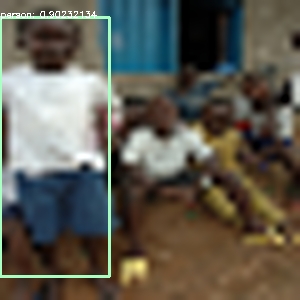}
      &
        \includegraphics[width=.18\textwidth]{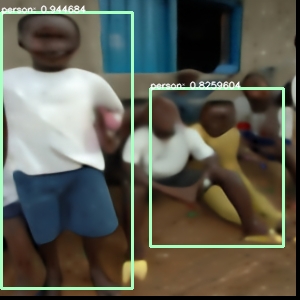}
      &
        \includegraphics[width=.18\textwidth]{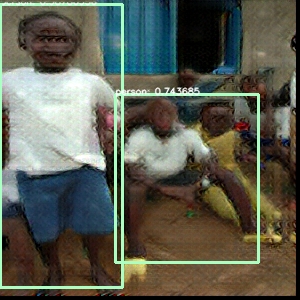}\\
      {\scriptsize (a) HR+Blur}
      &{\scriptsize (b) LR}
      &{\scriptsize (c) Bicubic}
      &{\scriptsize (d) SR-FT+}
      &{\scriptsize (e) TDSR}\\
    \end{tabular}
    \caption{Sample results on blur images for $4\times$ and
      $8\times$. Zoom in to
      see detection labels and scores.}
    \label{fig:blur1}
  \end{center}
\end{figure}

\begin{figure}[!t]
  \begin{center}
    \begin{tabular}[c]{cccccc}
      \includegraphics[width=.18\textwidth]{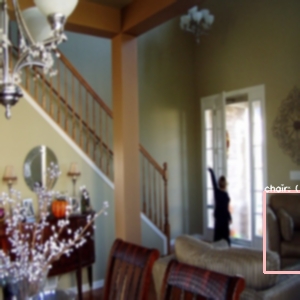}
      &
        \includegraphics[width=.18\textwidth]{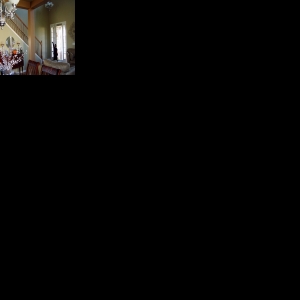}
      &
        \includegraphics[width=.18\textwidth]{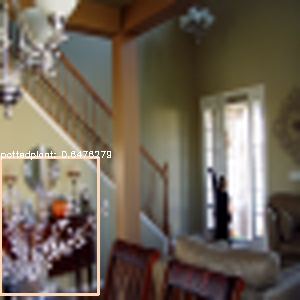}
      &
        \includegraphics[width=.18\textwidth]{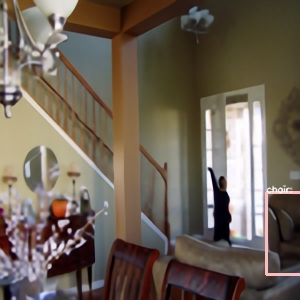}
      &
        \includegraphics[width=.18\textwidth]{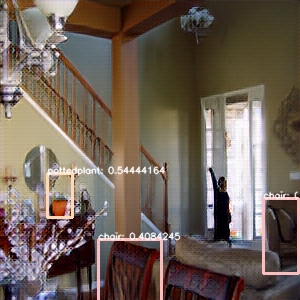}\\
        
        &
        \includegraphics[width=.18\textwidth]{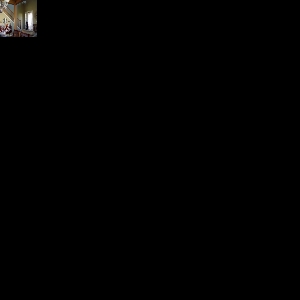}
      &
        \includegraphics[width=.18\textwidth]{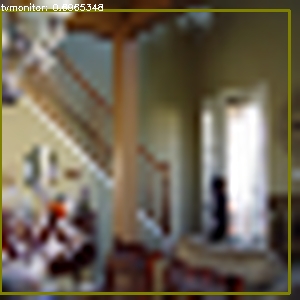}
      &
        \includegraphics[width=.18\textwidth]{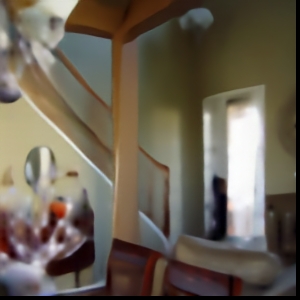}
      &
        \includegraphics[width=.18\textwidth]{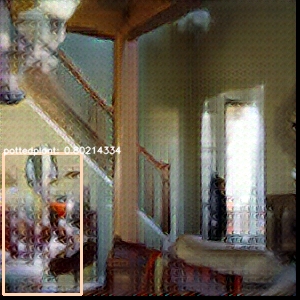}\\

      \includegraphics[width=.18\textwidth]{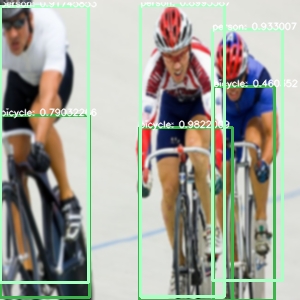}
      &
        \includegraphics[width=.18\textwidth]{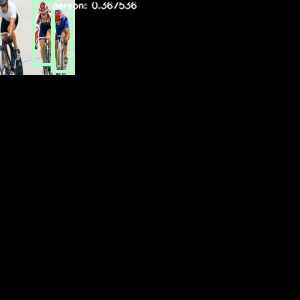}
      &
        \includegraphics[width=.18\textwidth]{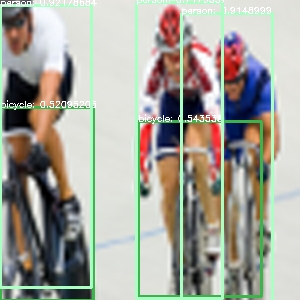}
      &
        \includegraphics[width=.18\textwidth]{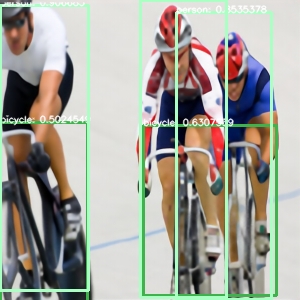}
      &
        \includegraphics[width=.18\textwidth]{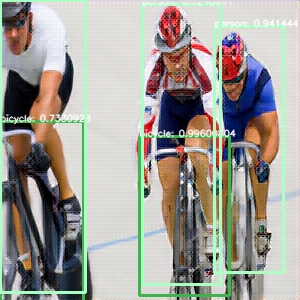}\\
        
        &
        \includegraphics[width=.18\textwidth]{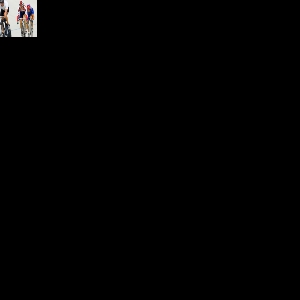}
      &
        \includegraphics[width=.18\textwidth]{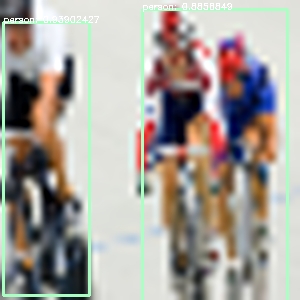}
      &
        \includegraphics[width=.18\textwidth]{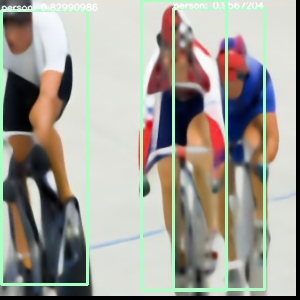}
      &
        \includegraphics[width=.18\textwidth]{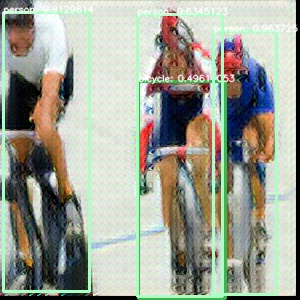}\\
        
        \includegraphics[width=.18\textwidth]{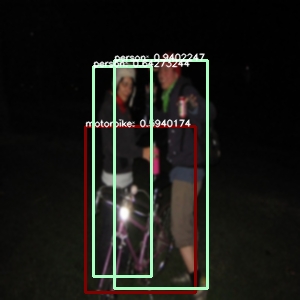}
      &
        \includegraphics[width=.18\textwidth]{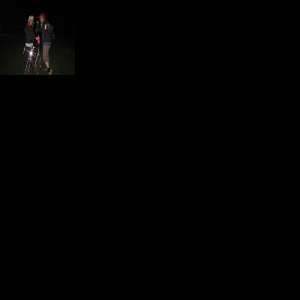}
      &
        \includegraphics[width=.18\textwidth]{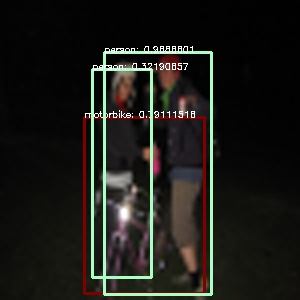}
      &
        \includegraphics[width=.18\textwidth]{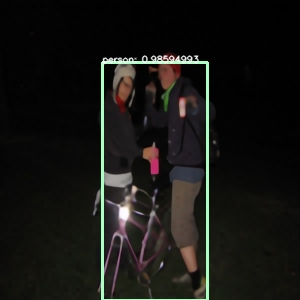}
      &
        \includegraphics[width=.18\textwidth]{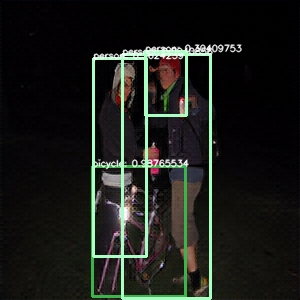}\\
        
        &
        \includegraphics[width=.18\textwidth]{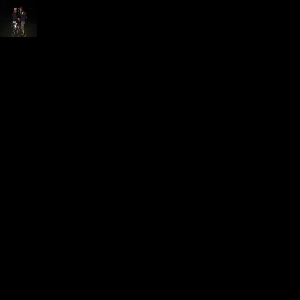}
      &
        \includegraphics[width=.18\textwidth]{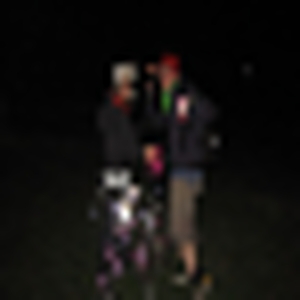}
      &
        \includegraphics[width=.18\textwidth]{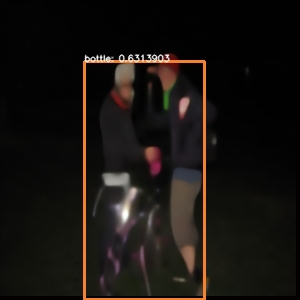}
      &
        \includegraphics[width=.18\textwidth]{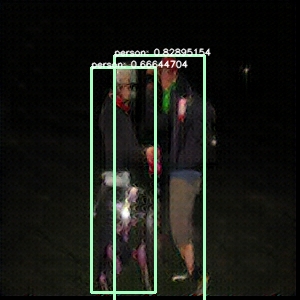}\\
        
         \includegraphics[width=.18\textwidth]{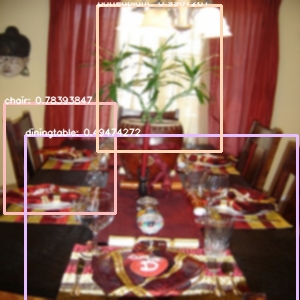}
      &
        \includegraphics[width=.18\textwidth]{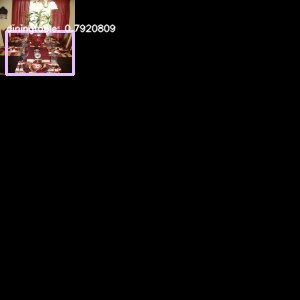}
      &
        \includegraphics[width=.18\textwidth]{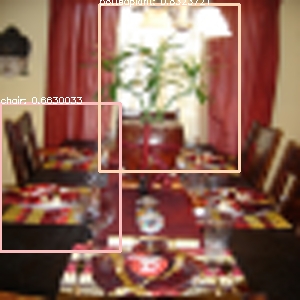}
      &
        \includegraphics[width=.18\textwidth]{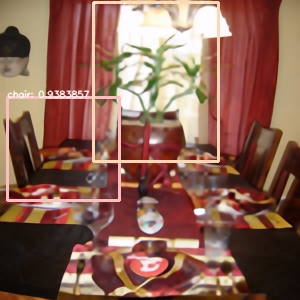}
      &
        \includegraphics[width=.18\textwidth]{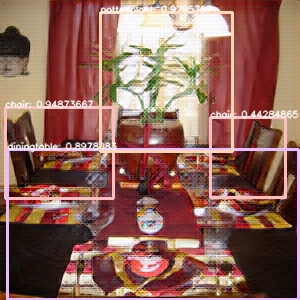}\\
        
        &
        \includegraphics[width=.18\textwidth]{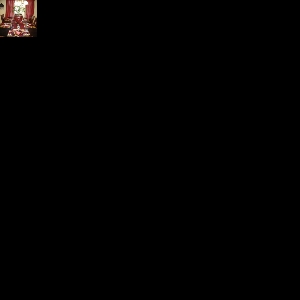}
      &
        \includegraphics[width=.18\textwidth]{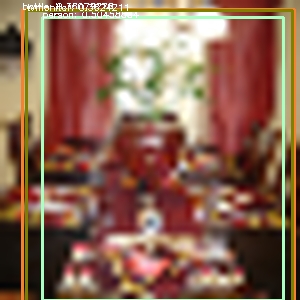}
      &
        \includegraphics[width=.18\textwidth]{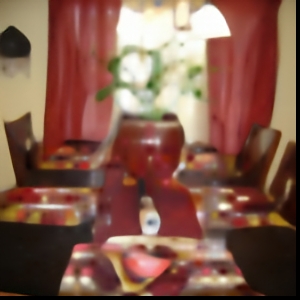}
      &
        \includegraphics[width=.18\textwidth]{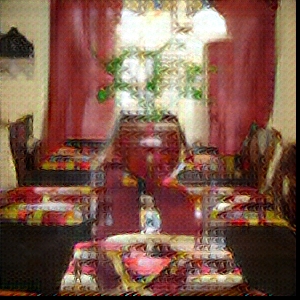}\\
      {\scriptsize (a) HR}
      &{\scriptsize (b) LR}
      &{\scriptsize (c) Bicubic}
      &{\scriptsize (d) SR-FT+}
      &{\scriptsize (e) TDSR}\\
    \end{tabular}
    \caption{Sample results on blur images for $4\times$ and
      $8\times$. Zoom in to
      see detection labels and scores.}
    \label{fig:blur2}
  \end{center}
\end{figure}

\begin{figure}[!t]
  \begin{center}
    \begin{tabular}[c]{ccccc}
      \includegraphics[width=.18\textwidth]{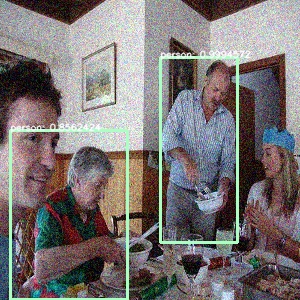}
      &
        \includegraphics[width=.18\textwidth]{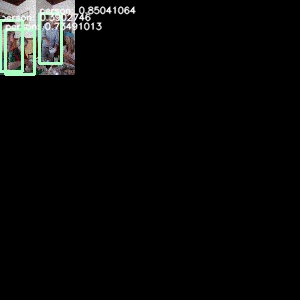}
      &
        \includegraphics[width=.18\textwidth]{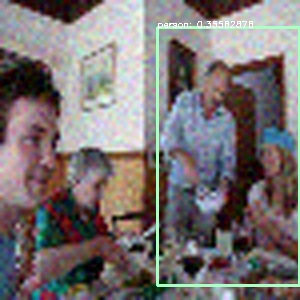}
        &
        \includegraphics[width=.18\textwidth]{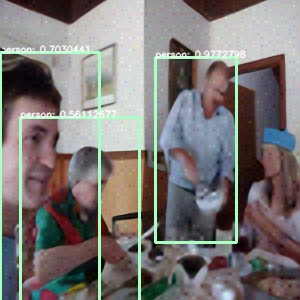}
      &
        \includegraphics[width=.18\textwidth]{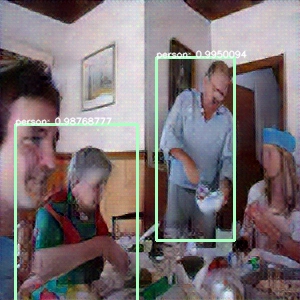}\\
        
        &
        \includegraphics[width=.18\textwidth]{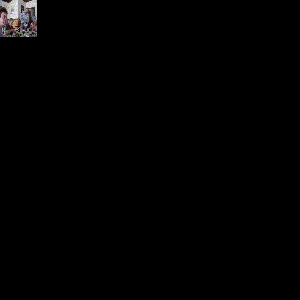}
      &
        \includegraphics[width=.18\textwidth]{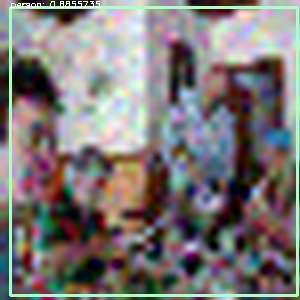}
        &
        \includegraphics[width=.18\textwidth]{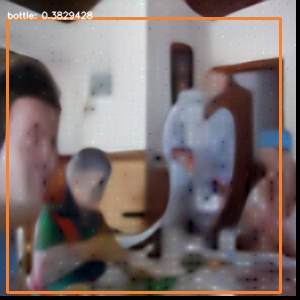}
      &
        \includegraphics[width=.18\textwidth]{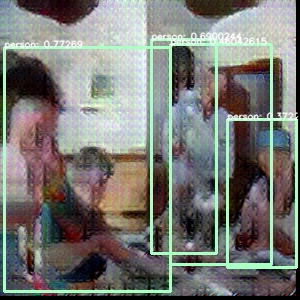}\\

      \includegraphics[width=.18\textwidth]{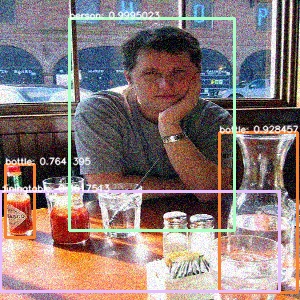}
      &
        \includegraphics[width=.18\textwidth]{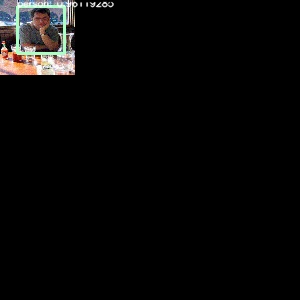}
      &
        \includegraphics[width=.18\textwidth]{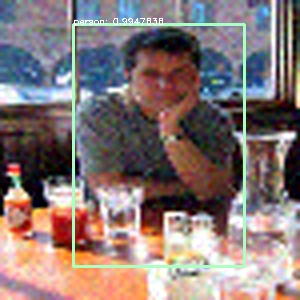}
      &
        \includegraphics[width=.18\textwidth]{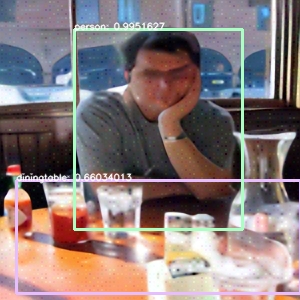}
      &
        \includegraphics[width=.18\textwidth]{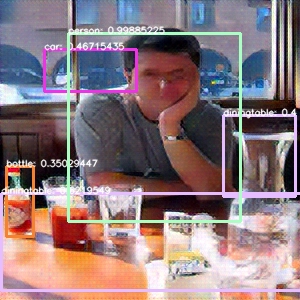}\\
        
        &
        \includegraphics[width=.18\textwidth]{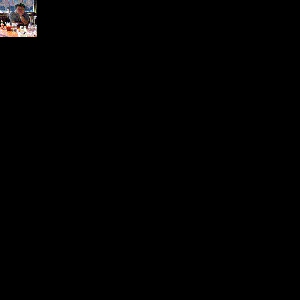}
      &
        \includegraphics[width=.18\textwidth]{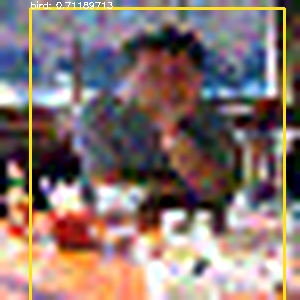}
      &
        \includegraphics[width=.18\textwidth]{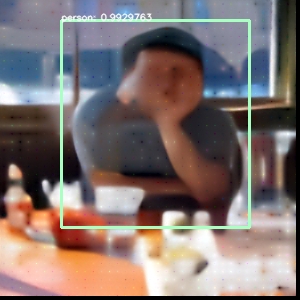}
      &
        \includegraphics[width=.18\textwidth]{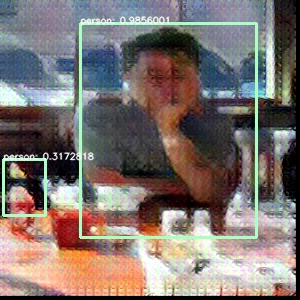}\\
        
        \includegraphics[width=.18\textwidth]{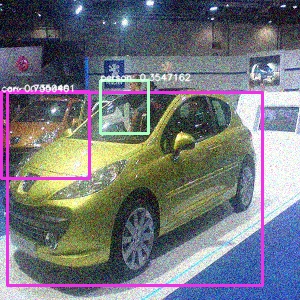}
      &
        \includegraphics[width=.18\textwidth]{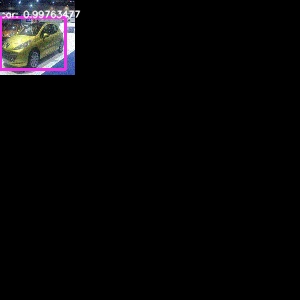}
      &
        \includegraphics[width=.18\textwidth]{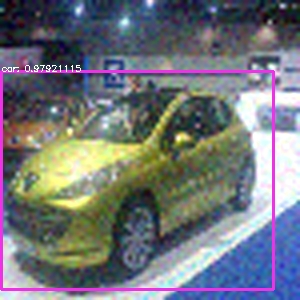}
      &
        \includegraphics[width=.18\textwidth]{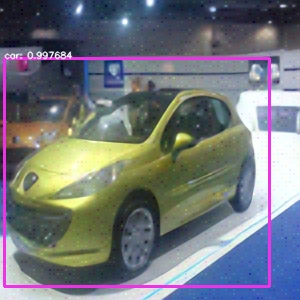}
      &
        \includegraphics[width=.18\textwidth]{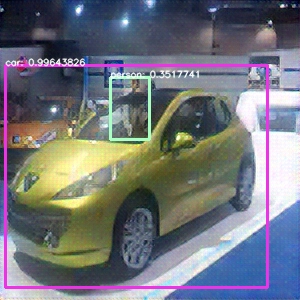}\\
        
        &
        \includegraphics[width=.18\textwidth]{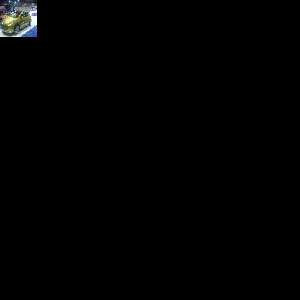}
      &
        \includegraphics[width=.18\textwidth]{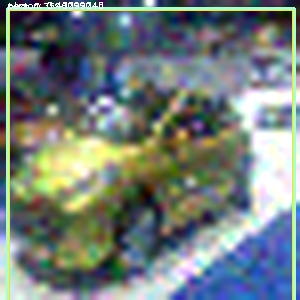}
      &
        \includegraphics[width=.18\textwidth]{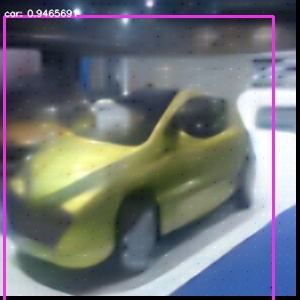}
      &
        \includegraphics[width=.18\textwidth]{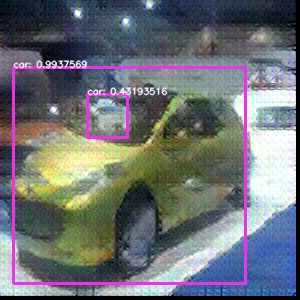}\\
        
          \includegraphics[width=.18\textwidth]{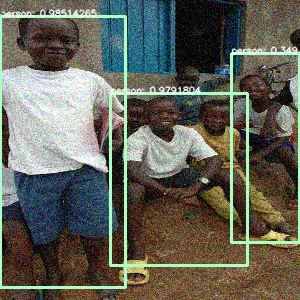}
      &
        \includegraphics[width=.18\textwidth]{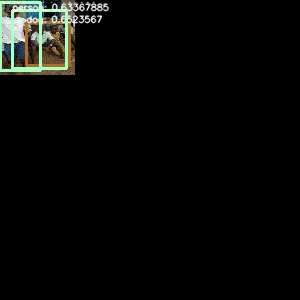}
      &
        \includegraphics[width=.18\textwidth]{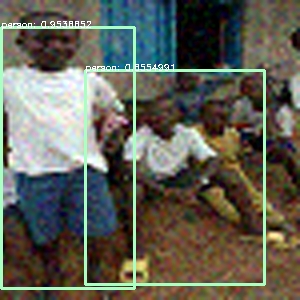}
      &
        \includegraphics[width=.18\textwidth]{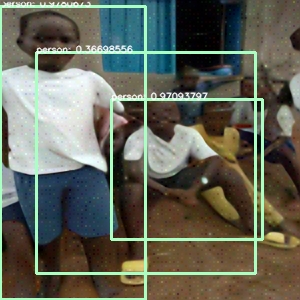}
      &
        \includegraphics[width=.18\textwidth]{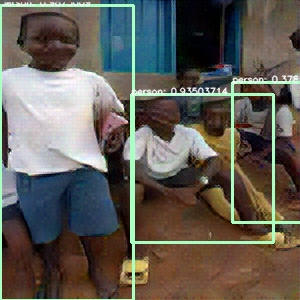}\\
        
        &
        \includegraphics[width=.18\textwidth]{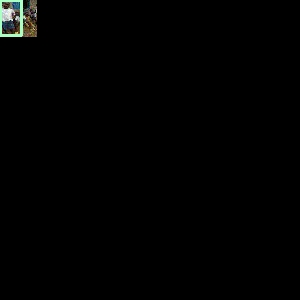}
      &
        \includegraphics[width=.18\textwidth]{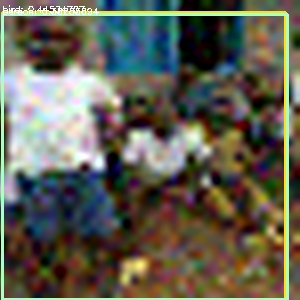}
      &
        \includegraphics[width=.18\textwidth]{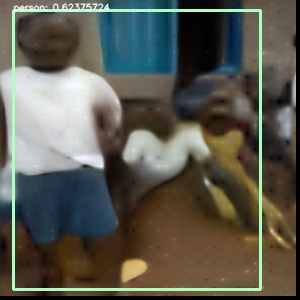}
      &
        \includegraphics[width=.18\textwidth]{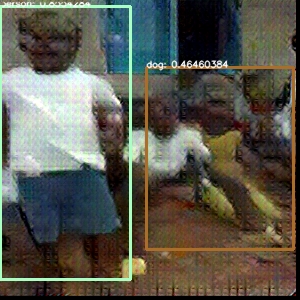}\\
      {\scriptsize (a) HR+Noise}
      &{\scriptsize (b) LR}
      &{\scriptsize (c) Bicubic}
      &{\scriptsize (d) SR-FT+}
      &{\scriptsize (e) TDSR}\\
    \end{tabular}
    \caption{Sample results on noise images for $4\times$ and
      $8\times$. Zoom in to
      see detection labels and scores.}
    \label{fig:noise1}
  \end{center}
\end{figure}

\begin{figure}[!t]
  \begin{center}
    \begin{tabular}[c]{cccccc}
      \includegraphics[width=.18\textwidth]{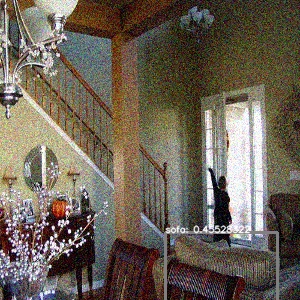}
      &
        \includegraphics[width=.18\textwidth]{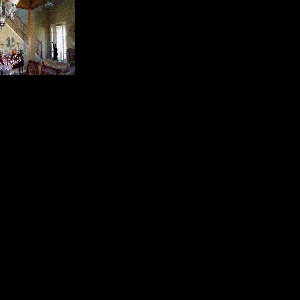}
      &
        \includegraphics[width=.18\textwidth]{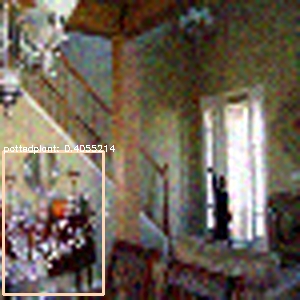}
      &
        \includegraphics[width=.18\textwidth]{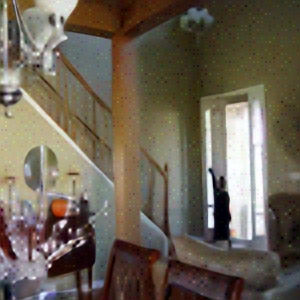}
      &
        \includegraphics[width=.18\textwidth]{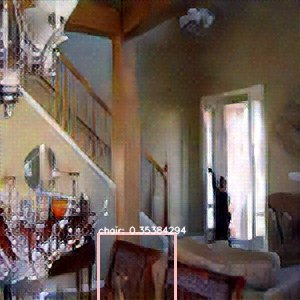}\\
        
        &
        \includegraphics[width=.18\textwidth]{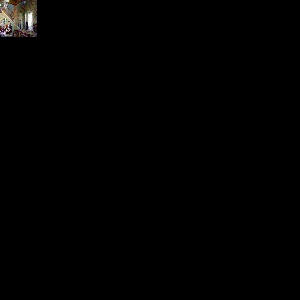}
      &
        \includegraphics[width=.18\textwidth]{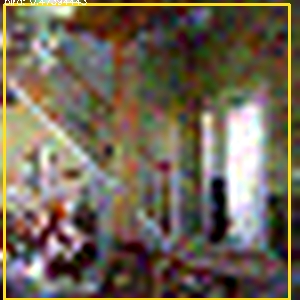}
      &
        \includegraphics[width=.18\textwidth]{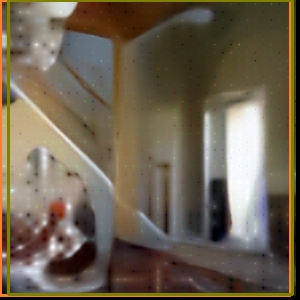}
      &
        \includegraphics[width=.18\textwidth]{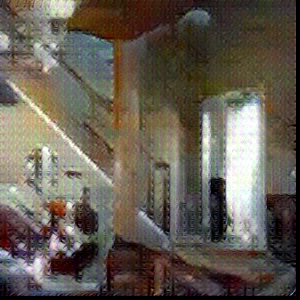}\\

      \includegraphics[width=.18\textwidth]{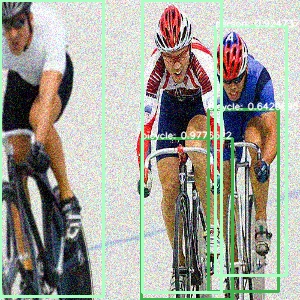}
      &
        \includegraphics[width=.18\textwidth]{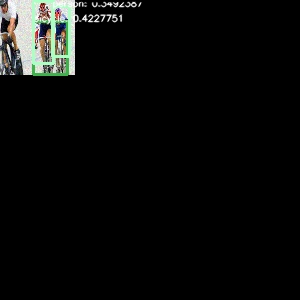}
      &
        \includegraphics[width=.18\textwidth]{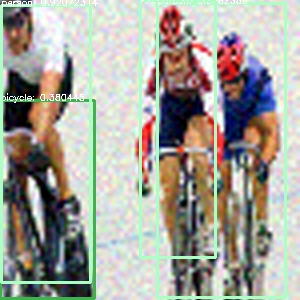}
      &
        \includegraphics[width=.18\textwidth]{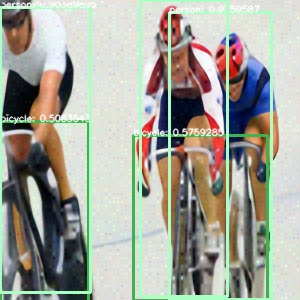}
      &
        \includegraphics[width=.18\textwidth]{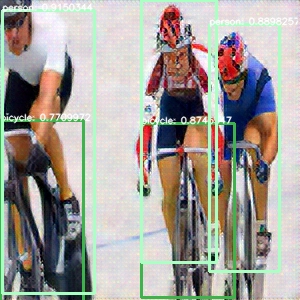}\\
        
        &
        \includegraphics[width=.18\textwidth]{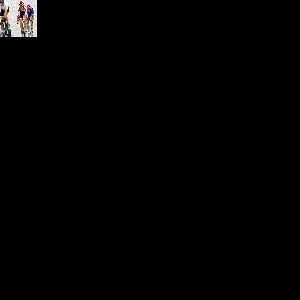}
      &
        \includegraphics[width=.18\textwidth]{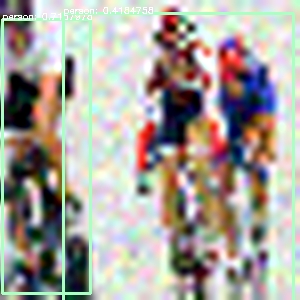}
      &
        \includegraphics[width=.18\textwidth]{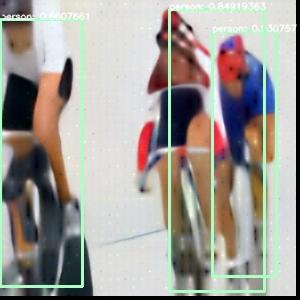}
      &
        \includegraphics[width=.18\textwidth]{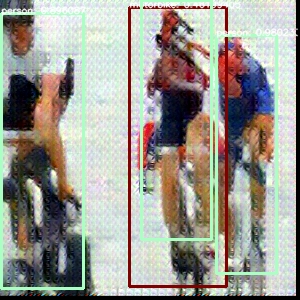}\\
        
        \includegraphics[width=.18\textwidth]{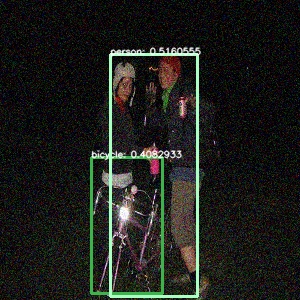}
      &
        \includegraphics[width=.18\textwidth]{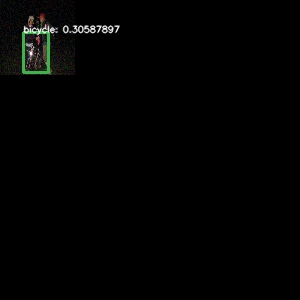}
      &
        \includegraphics[width=.18\textwidth]{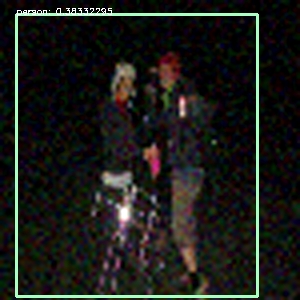}
      &
        \includegraphics[width=.18\textwidth]{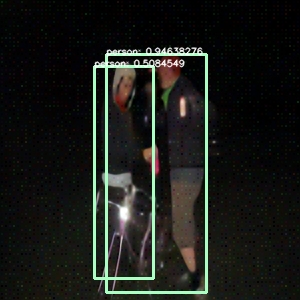}
      &
        \includegraphics[width=.18\textwidth]{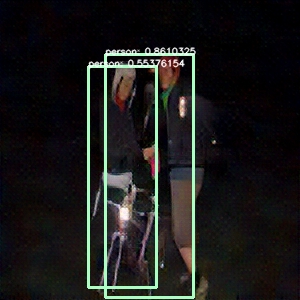}\\
        
        &
        \includegraphics[width=.18\textwidth]{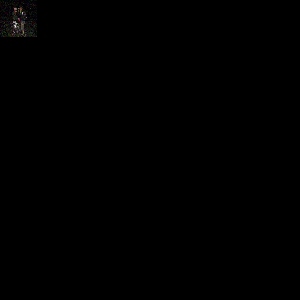}
      &
        \includegraphics[width=.18\textwidth]{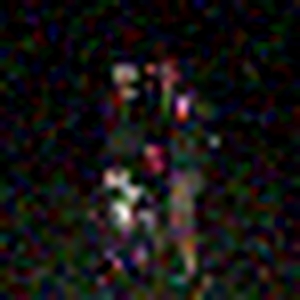}
      &
        \includegraphics[width=.18\textwidth]{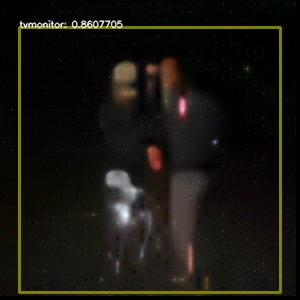}
      &
        \includegraphics[width=.18\textwidth]{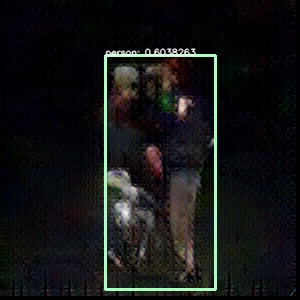}\\
        
         \includegraphics[width=.18\textwidth]{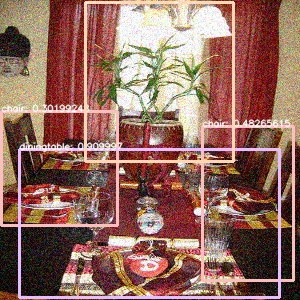}
      &
        \includegraphics[width=.18\textwidth]{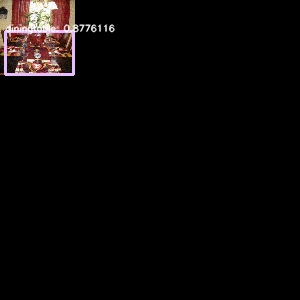}
      &
        \includegraphics[width=.18\textwidth]{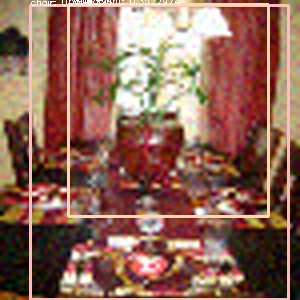}
      &
        \includegraphics[width=.18\textwidth]{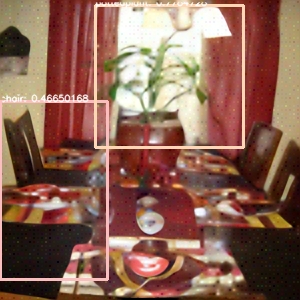}
      &
        \includegraphics[width=.18\textwidth]{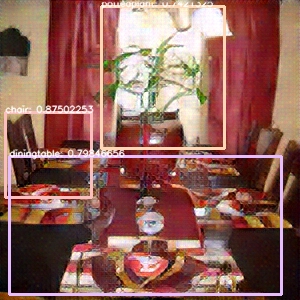}\\
        
        &
        \includegraphics[width=.18\textwidth]{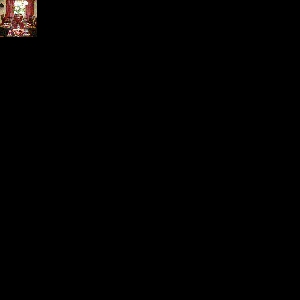}
      &
        \includegraphics[width=.18\textwidth]{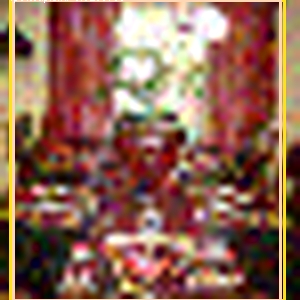}
      &
        \includegraphics[width=.18\textwidth]{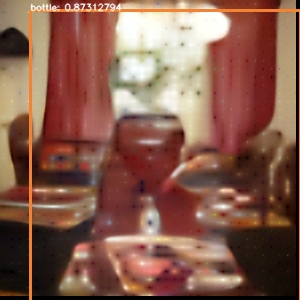}
      &
        \includegraphics[width=.18\textwidth]{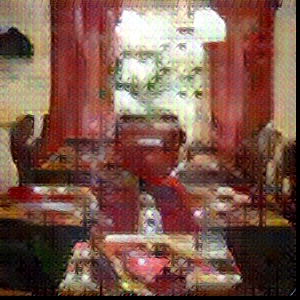}\\
      {\scriptsize (a) HR}
      &{\scriptsize (b) LR}
      &{\scriptsize (c) Bicubic}
      &{\scriptsize (d) SR-FT+}
      &{\scriptsize (e) TDSR}\\
    \end{tabular}
    \caption{Sample results on noise images for $4\times$ and
      $8\times$. Zoom in to
      see detection labels and scores.}
    \label{fig:noise2}
  \end{center}
\end{figure}

We provide more detection results in Fig.~\ref{fig:natural1},~\ref{fig:natural2},~\ref{fig:blur1},~\ref{fig:blur2},~\ref{fig:noise1},~and~\ref{fig:noise2}.

\clearpage
\bibliographystyle{splncs}
\bibliography{egbib}
\end{document}